\documentclass[lettersize,journal]{IEEEtran}
\usepackage{amsmath,amsfonts}
\usepackage{algorithmic}
\usepackage{algorithm}
\usepackage{array}
\usepackage[caption=false,font=normalsize,labelfont=sf,textfont=sf]{subfig}
\usepackage{textcomp}
\usepackage{stfloats}
\usepackage{url}
\usepackage{verbatim}
\usepackage{graphicx}
\usepackage{cite}
\usepackage{booktabs}
\usepackage{multirow}
\usepackage{threeparttable}
\usepackage[hidelinks]{hyperref}
\begin{document}

\title{Real-World Perception for Autonomous Driving in Adverse Weather: Enhancing Standard Detectors via Foundation-Guided Auto-Annotation}

\author{Sepideh Gohari, Goodarz Mehr, Azim Eskandarian,~\IEEEmembership{Fellow,~IEEE}
        % <-this % stops a space
\thanks{Sepideh Gohari, Goodarz Mehr, and Azim Eskandarian are with the Autonomous Robots and Vehicles Laboratory (ARVL), College of Engineering, Virginia Commonwealth University (VCU), Richmond, Virginia, United States. E-mail: \{goharis, mehrg, eskandariana\}@vcu.edu (\textit{Corresponding author: Sepideh Gohari})}}

\markboth{IEEE Transactions on Intelligent Transportation Systems}%
{Gohari \MakeLowercase{\textit{et al.}}: Real-World Perception for Autonomous Driving in Adverse Weather: Enhancing Standard Detectors via Foundation-Guided Auto-Annotation}


\maketitle

\begin{abstract}
Standard deployment-ready object detectors for autonomous vehicles degrade in adverse weather and lighting conditions without being trained on extensive domain-specific data. While large-scale vision foundation models offer robust zero-shot generalization, their high computational cost makes them impractical for real-time deployment. To bridge this gap, we propose a foundation-guided auto-annotation pipeline that enhances standard detectors without architectural changes. We first benchmark three distinct models, YOLOv8, Co-DETR, and SAM3, on our custom real-world driving dataset spanning 25 unique operational scenarios across various route, weather, and lighting conditions. Based on our analysis, SAM3 demonstrates superior accuracy and resilience across all scenarios. Thus, we deploy it as an offline auto-annotator to generate pseudo-labels on the unannotated subset of our dataset. Fine-tuning the baseline YOLOv8 on these annotations yields a 16.04\% higher overall mean Average Precision (mAP) and improves cross-environmental stability compared to the baseline model, highlighted by a 32.73\% and 28.65\% mAP increase in Residential Direct Sunlight and Highway Fog, respectively. These results demonstrate that standard detectors can achieve environmental resilience without the need for extensive manual annotation or architectural modifications.
\end{abstract}

\begin{IEEEkeywords}
Adverse weather perception, object detection, real-world dataset, auto-annotation, pseudo-labeling, autonomous driving.

% Real-world adverse weather dataset, pseudo-labeling, data-based approaches, computer vision, autonomous driving.

\end{IEEEkeywords}

\section{Introduction}
\label{sec:introduction_and_background}
\IEEEPARstart{A}{utonomous} vehicles have made significant progress in recent years, performing reliably within their designated operation areas. However, the unreliability of vision-based perception systems in adverse weather conditions hindered their expansion into broader environments \cite{yuan2026awd}. While modern 2D object detection algorithms, such as YOLO and its variants \cite{redmon2016you}, perform well under ideal weather and lighting conditions \cite{zou2023object}, their performance degrades when they are subjected to real-world physical noise such as rain and snow \cite{yuan2026awd}. Since these algorithms are based on convolutional neural networks (CNNs) and rely on localized receptive fields, they depend heavily on fine-grained textures and clear edges. In adverse weather and lighting conditions, physical noise degrades these visual cues by corrupting the high-frequency textures and fine structural geometries. Lacking broader spatial context, these localized features are no longer sufficient for reliable detection \cite{tran2025towards, vinciguerra2026clouded, patel2024comprehensive}. As a result, the number of misclassifications and missed detections increases as environmental conditions deteriorate \cite{pettersen2026robustness}. 

\begin{figure}[!t]
    \centering
    \includegraphics[width=\columnwidth]{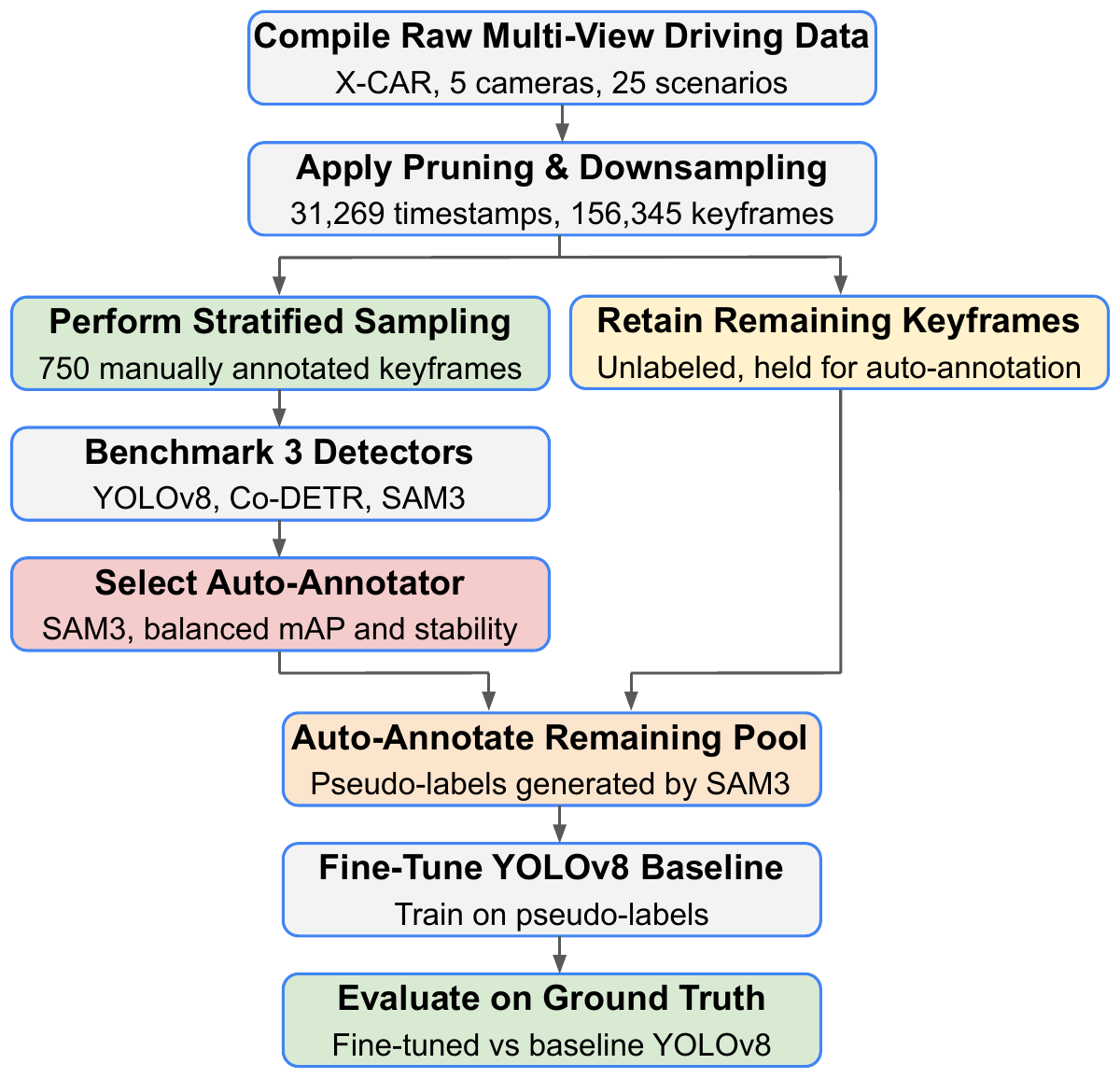}
    \caption{End-to-end framework of the proposed foundation-guided perception pipeline.}
    \label{fig:method_flowchart}
    \vspace{-10pt}
\end{figure}

Overcoming these limitations requires addressing challenges within both data curation and the model structure pipeline. On the data front, existing approaches face a dual bottleneck. Researchers frequently rely on synthetically generated adverse weather datasets to avoid manual labeling \cite{sakaridis2018semantic, halder2019physics}. However, such algorithmic augmentations cannot fully replicate the complex physical noise and sensor artifacts present in real-world data \cite{sakaridis2021acdc}. On the other hand, collecting real-world adverse weather data requires manual ground-truth annotation, which is prohibitively labor-intensive \cite{sakaridis2021acdc}. A solution to these problems is a framework that can exploit unannotated real-world driving data without relying on extensive manual effort.

On the model structure front, perception pipelines have increasingly adopted attention-based paradigms, ranging from Detection Transformers (DETR) \cite{carion2020end} to large-scale vision foundation models like the Segment Anything Model (SAM) \cite{kirillov2023segment} and its more recent variant, SAM3 \cite{carion2025sam}. Unlike traditional CNNs, these models replace localized receptive fields with local (windowed) and global attention. By capturing short- and long-range contextual relationships across the entire image, these models can infer object presence even when high-frequency textures are degraded by physical noise \cite{naseer2021intriguing}. Furthermore, pre-training on large-scale datasets equips them with rich spatial priors and robust zero-shot generalization across challenging environments \cite{wang2024exploring}. However, their large size and heavy computational requirements make direct deployment on autonomous vehicles for real-time inference impractical \cite{saha2025vision, lin2025deploying}.

A practical solution to this trade-off is offline knowledge transfer, which combines the resilience of foundation models to physical noise with the speed of standard detectors. In this framework, a large-scale foundation model acts as the offline pseudo-labeler of unannotated, visually challenging data \cite{radosavovic2018data, chen2018pseudo}. Fine-tuning lightweight detectors on these generated annotations enables them to capture complex environmental representations \cite{xu2023efficient}, which improves their performance in adverse conditions without modifying their architecture or increasing inference latency.

To implement this strategy, we propose the framework illustrated in Fig.~\ref{fig:method_flowchart}. We use a real-world driving dataset specifically collected by our research group for adverse weather perception, covering diverse weather, lighting, and route conditions \cite{du2024development}. On a manually annotated subset, we evaluate three architecturally distinct candidate models: YOLOv8 (CNN-based) \cite{yolov8_ultralytics}, Co-DETR (transformer-based) \cite{zong2023detrs}, and SAM3 (vision foundation) \cite{carion2025sam}. Among these candidates, SAM3 achieves optimal overall performance with the highest mAP and low standard deviation across all scenarios, confirming its viability as an offline auto-annotator. We then use SAM3 to generate pseudo-labels for the unannotated keyframes of the dataset and fine-tune baseline YOLOv8 on this data. This pipeline substantially improves YOLOv8's detection accuracy in adverse weather and lighting conditions while preserving its original architecture. 

In summary, our main contributions are as follows:
\begin{itemize}
    \item We present a comparison of three architecturally distinct detectors: YOLOv8, Co-DETR, and SAM3. We evaluate these models across our custom real-world driving dataset \cite{du2024development}, covering 25 combinations of route (e.g., Campus, Highway), weather (e.g., Rain, Fog), and lighting (e.g., Direct sunlight, Low light) conditions.
    \item We implement a foundation-guided auto-annotation pipeline that uses SAM3 to generate offline pseudo-labels for unannotated adverse weather driving data, eliminating the need for costly manual annotation.
    \item We show that fine-tuning baseline YOLOv8 on these pseudo-labels increases overall mAP in adverse conditions by 16.04\% (from a 34.19\% baseline mAP to 50.23\%), with peak gains of 32.73\% in Residential Direct Sunlight and 28.65\%  in Highway Fog, all without modifying YOLOv8's original architecture.
    % \item We show that fine-tuning baseline YOLOv8 on these pseudo-labels increases overall accuracy in adverse conditions by 16.04\% mAP (from 34.19\% mAP baseline to 50.23\% mAP fine-tuned), with peak gains of 32.73\% mAP in Residential Direct Sunlight and 28.65\% mAP in Highway Fog, all without modifying its original architecture.
\end{itemize}

\section{Methodology}
\label{sec:methodology}
This section presents our end-to-end pipeline, starting from dataset curation and stratified ground-truth annotation to foundation-guided auto-annotation and fine-tuning. An overview of the framework is shown in Fig.~\ref{fig:method_flowchart}.
% This section presents our end-to-end pipeline, progressing from dataset curation and stratified ground-truth annotation to candidate model selection and foundation-guided auto-annotation with fine-tuning. An overview of the framework is shown in Fig.~\ref{fig:method_flowchart}. 

\subsection{Dataset Overview and Adverse Conditions Taxonomy}
The dataset used in this work was specifically designed by our research group \cite{du2024development} as a small-scale, controlled benchmark for driving under adverse weather and lighting conditions. This dataset was collected using the X-CAR research platform \cite{mehr2022x}, which is equipped with five surround-view cameras (2880 $\times$ 1860 resolution). Consequently, each recorded timestamp in the dataset consists of five high-definition surround-view images. The dataset spans four distinct route topologies (Campus, Highway, Residential, Rural) with six independent conditions, comprising three weather domains (Fog, Rain, Snow) and three lighting domains (Direct Sunlight, Low Light, No Light). Including an additional Sufficient Light for the Campus route, this structured taxonomy yields 25 unique operational scenarios.

\subsection{Dataset Pruning and Keyframe Extraction}
To adapt this raw dataset for our specific evaluation and auto-annotation pipelines, we implemented a temporal pruning process and eliminated out-of-domain data. While the original continuous collection comprised 1,218,990 raw frames across the five surround-view cameras (representing 243,798 distinct timestamps), off-route transit periods at the beginning and end of each recording sequence (e.g., initial startup or final parking maneuvers) were removed to ensure the data strictly reflected the target operational route. This boundary isolation reduced the dataset to 849,795 active, in-domain driving frames (169,959 timestamps per camera).

The raw sensor data was originally captured at a frequency of 10 Hz. Sampling at this rate introduces visual redundancy, as frames captured 0.1 seconds apart are nearly identical. To resolve this, we applied a temporal downsampling protocol and extracted keyframes at a rate of 2 Hz. This interval was selected to create an optimal balance, it eliminates consecutive frame duplication while ensuring that scene transitions remain continuous.

This filtering, followed by the downsampling process, reduced the data into a pool of 31,269 timestamps, yielding a total multi-view collection of 156,345 keyframes across the five cameras. This processed dataset is then used for subsequent manual annotation and pseudo-labeling, with the keyframe distribution divided across the target routes: Campus (58,075), Residential (48,830), Rural (34,285), and Highway (15,155).

\begin{figure*}[!t]
    \centering
    \subfloat[]{\includegraphics[width=0.40\textwidth]{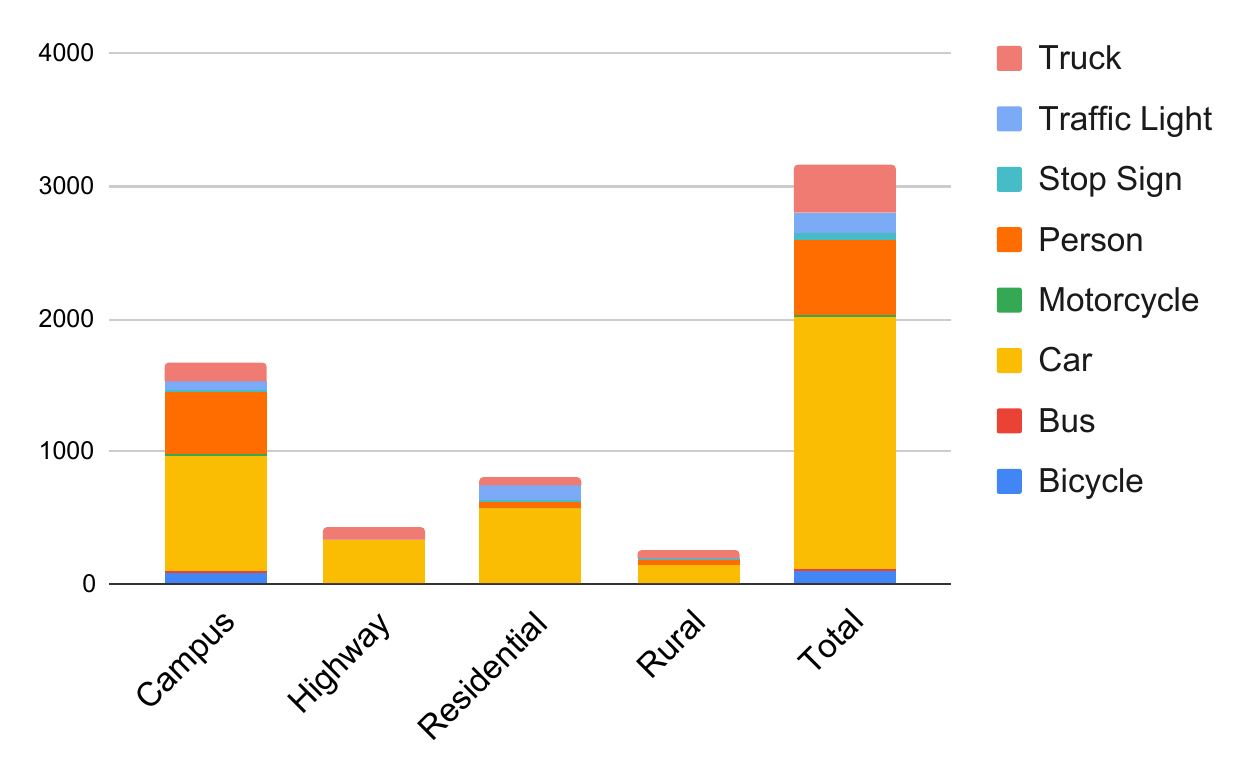}%
    \label{fig:fig_route_counts}}
    \hfil
    \subfloat[]{\includegraphics[width=0.40\textwidth]{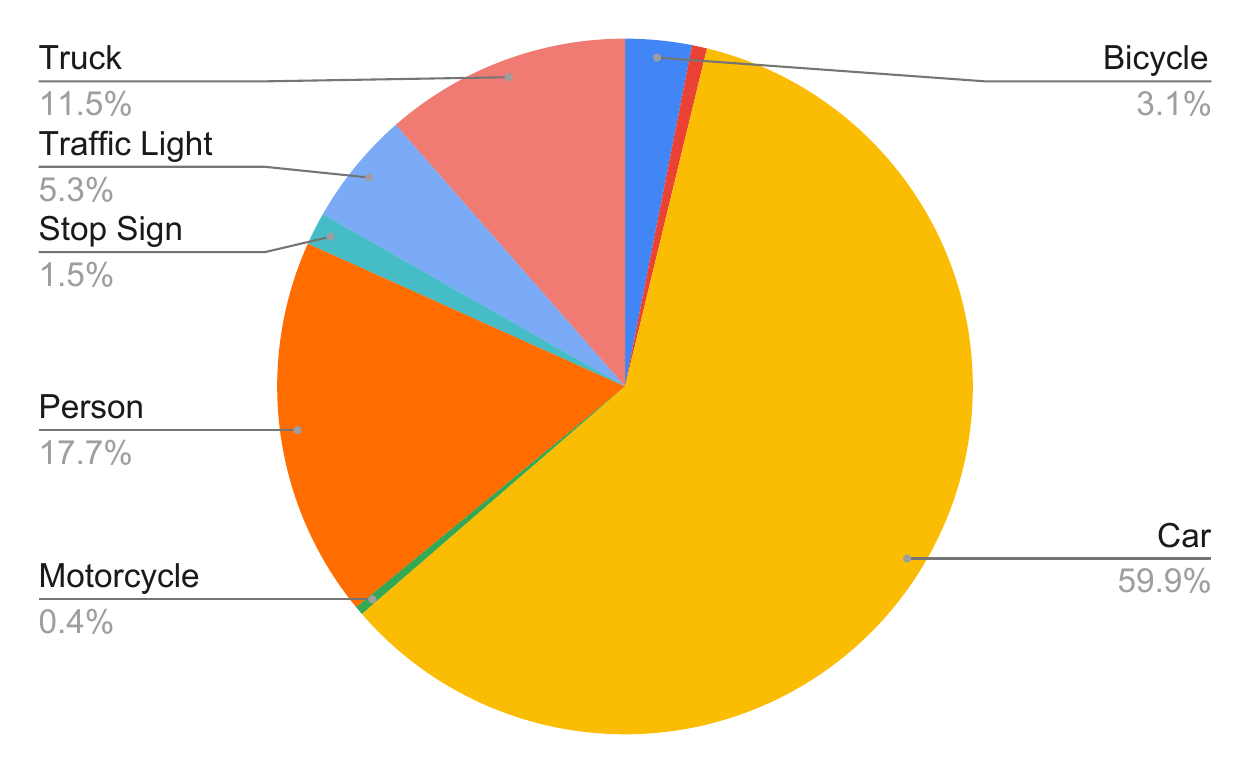}%
    \label{fig:fig_class_proportions}}
    \caption{Class distribution of the manually annotated ground-truth subset ($N=3,169$ total instances). (a) Instance count for each object class broken down by route type (Campus, Highway, Residential, Rural) and total count. (b) Overall percentage distribution of object classes across the ground-truth.
    }
    \label{fig:fig_dataset_distribution}
    \vspace{-10pt}
\end{figure*}
\subsection{Stratified Sampling and Ground-Truth Curation}
To construct an unbiased, reliable baseline for evaluating the candidate object detectors, we extracted a subset of the keyframe pool for manual ground-truth annotation. We established a sampling protocol to prevent environmental or spatial bias across the 25 operational scenarios. For each scenario, exactly 30 samples were selected. We distributed the sampling across the sensor suite by extracting six keyframes from each of the five surround-view cameras per scenario to guarantee complete 360° spatial representation. This structure yielded a balanced, multi-view evaluation set of 750 sampled keyframes (25 scenarios $\times$ 30 samples). The sample selection process was guided by two primary goals, maximizing visual diversity by enforcing a meaningful temporal distance between samples, and ensuring the presence of at least one target object of interest (e.g., a vehicle or pedestrian). However, satisfying both criteria while maintaining the rigid 30-sample criteria occasionally conflicted with real-world data distributions. Therefore, we relaxed both the temporal distance and object presence constraints in shorter recording sequences or sparsely populated environments. As a result, this sampling protocol yielded a small subset of samples containing no target objects. Retaining these background-only samples strengthens the evaluation by testing how well models handle empty scenes without triggering false-positive detections.

\subsection{Manual Annotation Protocol and Taxonomy}
After curating the subset of 750 samples, we manually annotated each sample using the Computer Vision Annotation Tool (CVAT) \cite{cvat} to construct our ground-truth. This ground-truth served as the benchmark to evaluate the performance of YOLOv8, Co-DETR, and SAM3 as three structurally distinct object detection frameworks under adverse conditions. To ensure a fair comparison, the ground-truth annotation categories had to directly match the supported output categories of all candidate models. SAM3 is an open-vocabulary foundation model with no constraints on its output categories. However, both YOLOv8 and Co-DETR are closed-vocabulary models and their detection output is restricted to the 80 standard categories of the Microsoft COCO dataset \cite{lin2014microsoft}. Consequently, our ground-truth categories were bounded by the COCO category set to maintain consistency across our candidate models. 

Among the standard COCO categories, only a subset of eight classes is directly relevant to driving perception tasks: car, truck, bus, bicycle, motorcycle, person, traffic light, and stop sign. To ensure the long-term utility and scalability of our curated ground-truth, we annotated this subset using 20 fine-grained classes. However, for the scope of this study, we merged these granular classes through a mapping layer to match the eight baseline COCO categories and excluded all non-target classes.

During annotation, we drew tight bounding boxes around each target instance to exclude unnecessary background pixels. To maintain consistent evaluation standards under adverse visibility, we annotated all identifiable objects where the bounding box dimensions were greater than 20 $\times$ 20 pixels, regardless of heavy occlusion or frame-edge truncation. For instances that were visually identifiable but fell below the 20 $\times$ 20 pixel threshold, we assigned them to the ignore category corresponding to their class. However, we removed all these ignore classes alongside the non-target categories prior to the evaluation process in this study. This annotation protocol ensures that small, ambiguous background objects do not artificially skew the precision and recall across our candidate detectors.

In total, this annotation procedure yielded 3,169 bounding box instances across our 750 samples. Fig.~\ref{fig:fig_route_counts} details the distribution of ground-truth categories across the four routes, as well as cumulative totals, and Fig.~\ref{fig:fig_class_proportions} shows the overall class proportions across the entire ground-truth. With an average density of 4.23 instances per frame, the dataset is dominated by standard vehicular actors (59.9\% cars) and vulnerable road users (17.7\% persons). This class imbalance reflects real-world driving conditions, enabling us to test how well detectors locate and classify objects across diverse operational domains.

\subsection{Evaluation Models}
We selected three object detection frameworks representing distinct architectural paradigms, namely a real-time detector (YOLOv8), a high-performance transformer (Co-DETR), and an open-vocabulary foundation model (SAM3). We evaluated these models against our manually annotated ground-truth to test their robustness under adverse weather and lighting conditions.

Our first candidate model is YOLOv8, a real-time, single-stage detector widely adopted for autonomous driving and perception tasks due to its recognized balance of computational efficiency and accuracy. Among its five scaled configurations (Nano, Small, Medium, Large, and Extra-Large), we selected the YOLOv8-Large (YOLOv8l) variant. While the Extra-Large (YOLOv8x) version offers a 1.0\% mAP gain on the COCO dataset, it introduces a 58.6\% latency penalty, increasing runtime from 9.06 to 14.37 ms. Selecting YOLOv8l gives us strong multi-class detection capabilities without over-parameterizing the model.

For our second candidate, we chose Co-DETR, a collaborative detection transformer that optimizes encoder learning through parallel auxiliary heads. To maintain consistency with the closed-vocabulary COCO classes used in the YOLOv8 setup, we selected a COCO-pretrained configuration featuring a Vision Transformer Large (ViT-L) backbone. This architecture achieves a detection accuracy of 65.9\% Average Precision on the COCO dataset. Deploying this high-capacity model establishes a performance upper bound, allowing us to compare performance against real-time detectors and open-vocabulary architectures.

Our third candidate model is SAM3, an open-vocabulary foundation model optimized for Promptable Concept Segmentation (PCS) using a text-conditioned, DETR-based detector. Despite its zero-shot setup, SAM3 achieves a competitive 56.4\% mAP on the COCO dataset. To ensure a fair comparison among the candidates, we evaluated SAM3 using text-only prompting without manual point guidance or image exemplars, extracting bounding boxes directly from its box prediction head. For resolving semantic ambiguity from broad open-vocabulary queries, we prompted the model with 14 fine-grained class names, which we subsequently mapped and merged back into the 8 baseline COCO driving categories during post-processing. Finally, we eliminated spatial redundancies caused by merging sub-classes, such as overlapping bounding boxes when merging "rider" and "pedestrian" into "person", by applying Non-Maximum Suppression (NMS) to generate clean, non-duplicated predictions.

\subsection{Auto-Annotation Framework and Fine-Tuning}
After evaluating the three candidate detectors under adverse weather and lighting conditions, we selected the architecture demonstrating the highest accuracy and environmental robustness as our offline auto-annotator. We then used this selected model to process unannotated keyframes across our entire dataset, generating a comprehensive set of pseudo-labels. To prevent data leakage and maintain a fair evaluation, we kept the manually annotated subset entirely separate from this auto-annotation pipeline, reserving it strictly as our ground-truth set for final evaluation.

Following this stage, we used the pool of keyframes annotated with these pseudo-labels to fine-tune the YOLOv8l model, adapting this general-domain detector for specialized perception under adverse weather and lighting conditions. With this setup, we evaluated how effectively knowledge transfers across models. Specifically, we examined whether a real-time, lightweight detector can learn from a large, high-capacity offline model to improve its detection performance under visually degraded conditions.

\section{Experiments and Discussion}
\label{sec:experiments_and_discussion}
\subsection{Implementation Details}
We ran all evaluations on an NVIDIA GeForce RTX 3090 GPU. To prevent dependency conflicts, we isolated each candidate model within its dedicated Docker environment. During inference, we set a unified confidence threshold of $\tau = 0.5$ across all models to balance precision and recall. This value prevents false positives caused by lower thresholds while avoiding the suppression of distant or occluded objects, such as vulnerable road users, at higher thresholds.

\subsection{Evaluation Metrics}
We evaluated model performance using the standard COCO protocol, adopting $\text{mAP}_{50:95}$ as our primary metric. For a given object class $c$ and Intersection over Union (IoU) threshold $i$, Average Precision ($AP$) integrates the Area Under the Curve (AUC) of the Precision-Recall distribution:

\begin{equation}
\label{AP_equation}
AP_{c, i} = \int_{0}^{1} P(R) \, dR
\end{equation}

The $\text{mAP}_{50:95}$ metric averages these AP values across all target classes and IoU thresholds ranging from 0.50 to 0.95.

\subsection{Candidate Detector Evaluation under Adverse Conditions}
Table~\ref{tab:master_results} presents the quantitative evaluation of YOLOv8l, Co-DETR (ViT-L), and SAM3 across varying weather and lighting conditions, measured via standard $\text{mAP}_{50:95}$ at a unified confidence threshold ($\tau = 0.5$).
\begin{table}
    \centering
    \begin{threeparttable}
    \caption{Performance comparison ($\text{mAP}_{50:95}$ (\%)) under varying weather and lighting conditions across four route types.}
    \label{tab:master_results}
    \begin{tabular}{l l c c c}
        \toprule
        \textbf{Environment} & \textbf{Condition} & \textbf{YOLOv8} & \textbf{CoDETR} & \textbf{SAM3} \\ 
        \midrule
        \multirow{7}{*}{Campus} 
            & Direct Sunlight & 29.71 & 48.44 & \textbf{53.47} \\
            & Fog             & 36.68 & 68.66 & \textbf{68.93} \\
            & Low Light       & 32.61 & \textbf{54.10} & 46.06 \\
            & No Light        & 37.19 & 38.63 & \textbf{51.72} \\
            & Rain            & 38.77 & \textbf{57.55} & 55.27 \\
            & Snow            & 30.88 & 52.10 & \textbf{55.58} \\
            & Sufficient Light\tnote{*}& 41.72 & \textbf{60.47} & 49.61 \\
        \midrule
        \multirow{6}{*}{Highway} 
            & Direct Sunlight & 34.86 & \textbf{54.64} & 50.13 \\
            & Fog             & 16.83 & 42.78 & \textbf{55.71} \\
            & Low Light       & 36.12 & \textbf{62.06} & 60.53 \\
            & No Light        & 12.88 & 25.62 & \textbf{26.82} \\
            & Rain            & 19.51 & \textbf{47.21} & 43.45 \\
            & Snow            & 25.74 & \textbf{53.28} & 48.40 \\
        \midrule
        \multirow{6}{*}{Residential} 
            & Direct Sunlight & 27.01 & \textbf{62.38} & 60.39 \\
            & Fog             & 26.40 & 42.39 & \textbf{45.43} \\
            & Low Light       & 49.98 & 54.97 & \textbf{68.45} \\
            & No Light        & 23.65 & 30.42 & \textbf{47.87} \\
            & Rain            & 44.78 & 56.63 & \textbf{65.55} \\
            & Snow            & 35.03 & \textbf{51.58} & 48.18 \\
        \midrule
        \multirow{6}{*}{Rural} 
            & Direct Sunlight & 48.80 & \textbf{76.36} & 72.36 \\
            & Fog             & 53.07 & \textbf{78.05} & 73.07 \\
            & Low Light       & 39.75 & 47.32 & \textbf{63.70} \\
            & No Light        & 28.23 & 29.04 & \textbf{53.70} \\
            & Rain            & 44.34 & \textbf{76.65} & 69.35 \\
            & Snow            & 47.68 & \textbf{66.29} & 63.19 \\
        \midrule
        \multirow{2}{*}{\textbf{Overall Summary}} 
            & Mean ($\mu$)      & 34.19 & 53.21 & \textbf{56.14} \\
            & STD ($\sigma$)     & \textbf{10.68} & 14.31 & 10.98 \\
        \bottomrule
    \end{tabular}
    \begin{tablenotes}[flushleft]
        \small
        \item[*] Serves strictly as Campus baseline reference and is excluded from the overall metrics ($\mu, \sigma$) to preserve cross-environmental symmetry.
    \end{tablenotes}
    \end{threeparttable}
    \vspace{-10pt}
\end{table}

Across all evaluated conditions, YOLOv8 consistently underperforms the transformer architectures, recording an overall mean of 34.19\% mAP compared to 53.21\% mAP for Co-DETR and 56.14\% mAP for SAM3. On average, YOLOv8 lags behind by roughly 20\% mAP, with the gap reaching a maximum difference of 38.88\% against SAM3 under Highway Fog (16.83\% vs. 55.71\% mAP). This gap stems from YOLO's dependence on local receptive fields. When adverse conditions such as fog or glare degrade fine texture cues, local features become unreliable. In contrast, Co-DETR and SAM3 use attention mechanisms, allowing them to infer bounding boxes using the broader spatial and semantic context.

Further differences are observed between the two transformer architectures under No Light conditions. Although performance drops across all models in darkness, SAM3 outperforms Co-DETR, particularly in No Light Rural settings (53.70\% vs. 29.04\% mAP). Since Co-DETR relies solely on visual features, it is vulnerable when underexposure eliminates image contrast. However, SAM3 uses text-conditioned semantic embeddings to guide visual cross-attention. Combined with its large-scale pre-training, this helps stabilize object localization even when visual features are degraded.

Additionally, all models experience a sharp accuracy drop under the Highway No Light scenario. In this setting, high driving speeds induce motion blur while a lack of proper lighting eliminates image contrast. Their combined effect yields the lowest scores across the dataset, 12.88\% for YOLO, 25.62\% for Co-DETR, and 26.82\% mAP for SAM3. This decline highlights a limitation of camera-only perception, showing that passive vision alone fails when adverse environmental conditions become severe.

Evaluating overall accuracy ($\mu$) and standard deviation ($\sigma$) across all conditions and routes reveals the differences in model stability. YOLO records the lowest standard deviation ($\sigma = 10.68$), which reflects its consistently low performance ($\mu = 34.19\%$). Between the two transformer-based models, Co-DETR achieves a high mean accuracy ($\mu = 53.21\%$) but exhibits a high standard deviation ($\sigma = 14.31$), showing that its predictions are sensitive to environmental changes. SAM3 offers the optimal balance, achieving the highest overall performance ($\mu = 56.14\%$) alongside high cross-environmental stability ($\sigma = 10.98$). Nevertheless, an overall average of 56.14\% mAP remains insufficient for safety-critical autonomous driving. These results indicate that even state-of-the-art vision models cannot overcome extreme environmental degradations on their own, reinforcing the need to integrate active sensors like LiDAR or Radar to improve safety and real-world reliability.

\subsection{Efficacy of Foundation-Guided Fine-Tuning}
Based on our initial evaluation, we selected SAM3 as our offline model due to its optimal balance of accuracy and stability across varying weather, lighting, and route conditions. We used this model to automatically annotate the unannotated keyframes, and applied these generated pseudo-annotations to fine-tune the baseline YOLO architecture. To evaluate the efficacy of this foundation-guided fine-tuning, we compared the baseline pre-trained YOLO against the fine-tuned model across all test scenarios, with overall metrics summarized in Table \ref{tab:finetune_yolo}.
\begin{table}
    \centering
    \caption{Overall Performance Summary: Baseline vs. Fine-Tuned YOLOv8.}
    \label{tab:finetune_yolo}
    \begin{tabular}{l c c}
        \toprule
        \textbf{Model Architecture} & \textbf{Mean mAP ($\mu$)} & \textbf{ STD ($\sigma$)} \\ 
        \midrule
        Baseline YOLO (Pre-trained) & 34.19 & 10.68 \\
        Fine-Tuned YOLO (SAM3-Guided) & \textbf{50.23} & \textbf{9.96} \\
        \midrule
        \textbf{Overall Improvement ($\Delta$)} & \textbf{+16.04} & \textbf{-0.72} \\
        \bottomrule
    \end{tabular}
    \vspace{-10pt}
\end{table}

Fine-tuning improved overall detection performance, raising the average mAP from 34.19\% to 50.23\%. Alongside this 16.04\% improvement, the standard deviation decreased from 10.68 to 9.96. The lower standard deviation indicates that the fine-tuned YOLO model inherited the cross-environmental stability of SAM3, yielding more consistent predictions across changing weather and lighting conditions.

\begin{figure}[!t]
    \centering
    \includegraphics[width=\columnwidth]{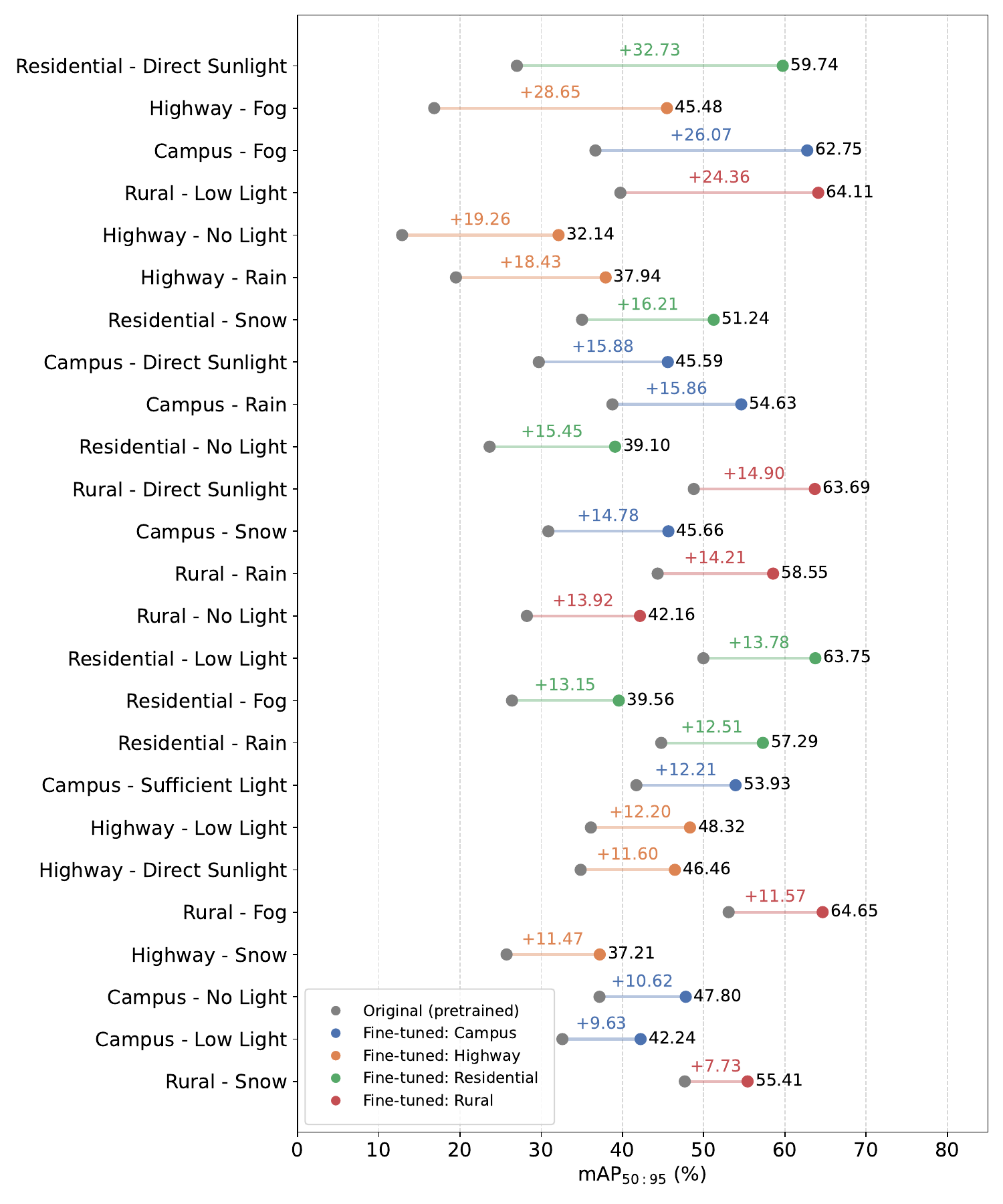}
    \caption{Performance improvements (mAP$_{50-95}$) of the SAM3-guided fine-tuned YOLOv8 model versus the pre-trained baseline, sorted by the magnitude of improvement. Grey markers indicate the baseline accuracy, while colored markers denote the fine-tuned model across all 25 scenarios.}
    \label{fig:finetuned_yolo}
    \vspace{-10pt}
\end{figure}

Fig.~\ref{fig:finetuned_yolo} illustrates performance changes across all 25 environmental conditions. The fine-tuning approach was particularly effective in adverse scenarios where the baseline model previously failed. Most notably, the fine-tuned model achieved major gains in conditions like Residential Direct Sunlight (+32.73\% mAP) and Highway Fog (+28.65\% mAP). These findings show that using a large offline foundation model for auto-annotation allows lightweight detectors to overcome feature representation limits without manual labeling effort, maintaining robust perception in adverse weather.

\subsection{Qualitative Comparison}
\begin{figure*}[t]
    \centering
    \small % Keeps header text readable and compact
    \setlength{\tabcolsep}{0.1pt} % Tightens space between columns
    \renewcommand{\arraystretch}{0.2} % Adjusts row spacing
    
    \begin{tabular}{c c c c c c}
      % Column Headers
      & \textbf{(a) Ground Truth} 
      & \textbf{(b) Baseline YOLOv8} 
      & \textbf{(c) Co-DETR} 
      & \textbf{(d) SAM3} 
      & \textbf{(e) Fine-Tuned YOLOv8} \\
    
      % Row 1: Residential Direct Sunlight
      \rotatebox[origin=l]{90}{\textbf{\footnotesize Res. Direct Light}} &
      \includegraphics[width=0.195\textwidth]{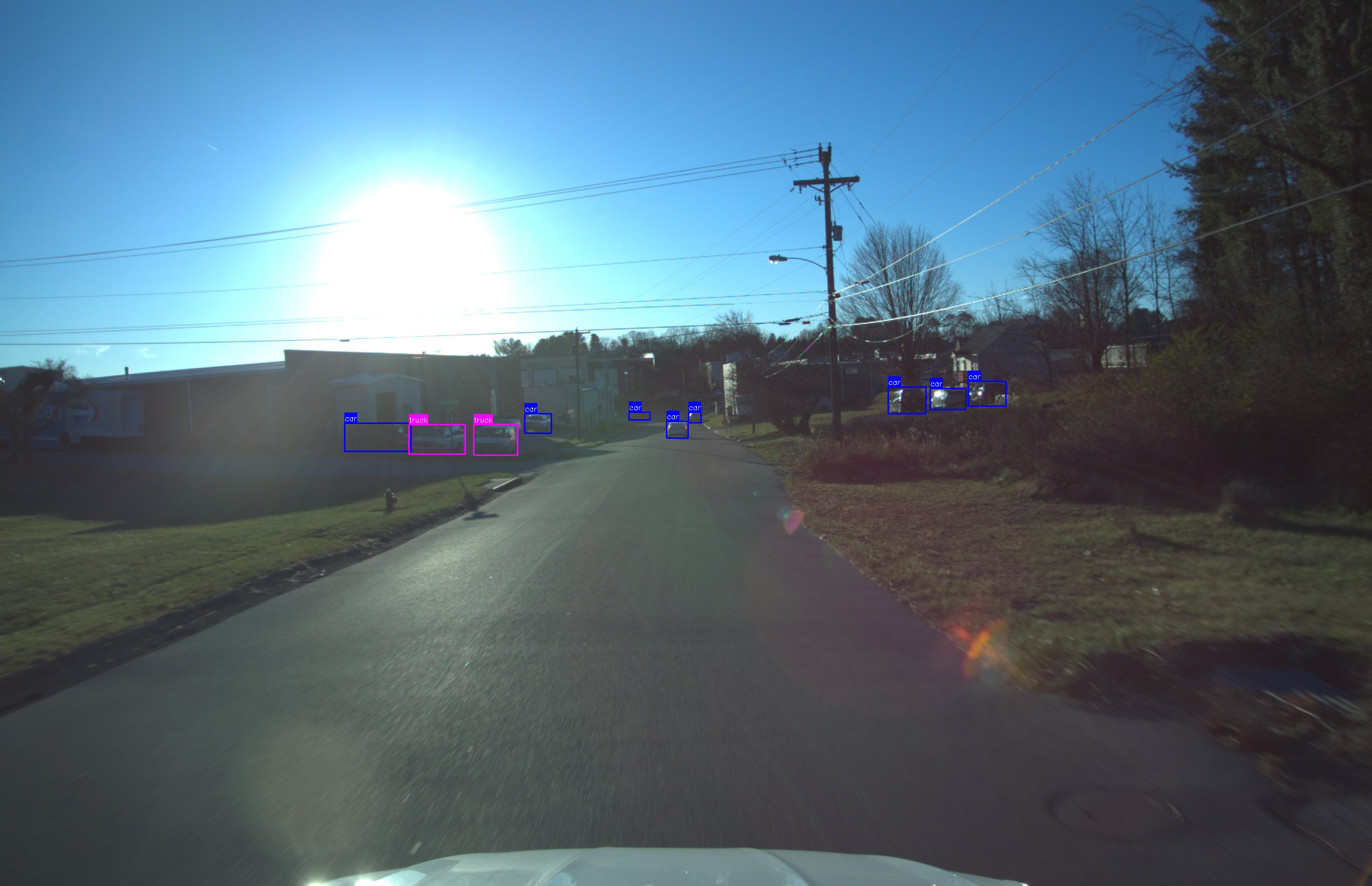} &
      \includegraphics[width=0.195\textwidth]{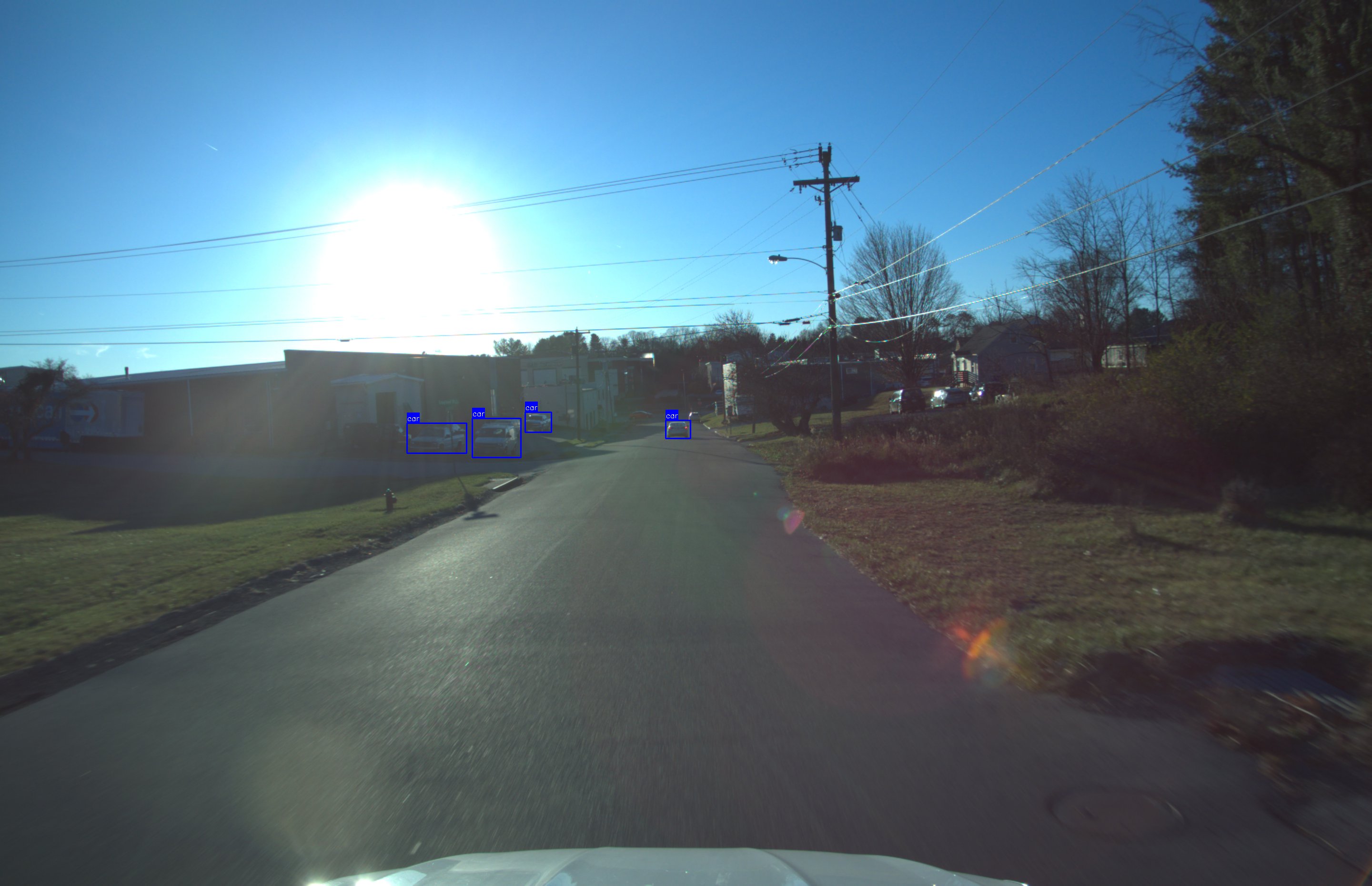} &
      \includegraphics[width=0.195\textwidth]{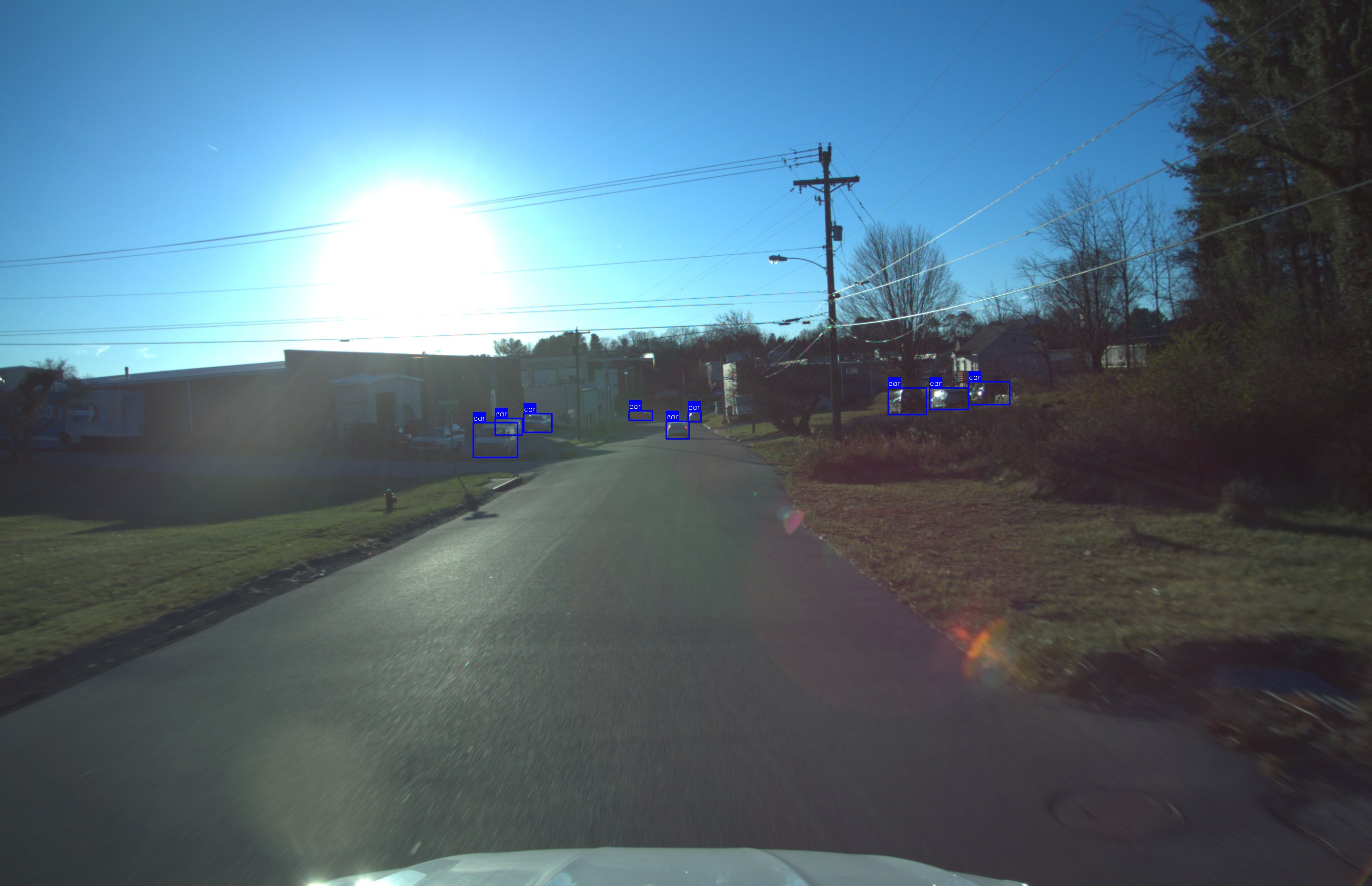} &
      \includegraphics[width=0.195\textwidth]{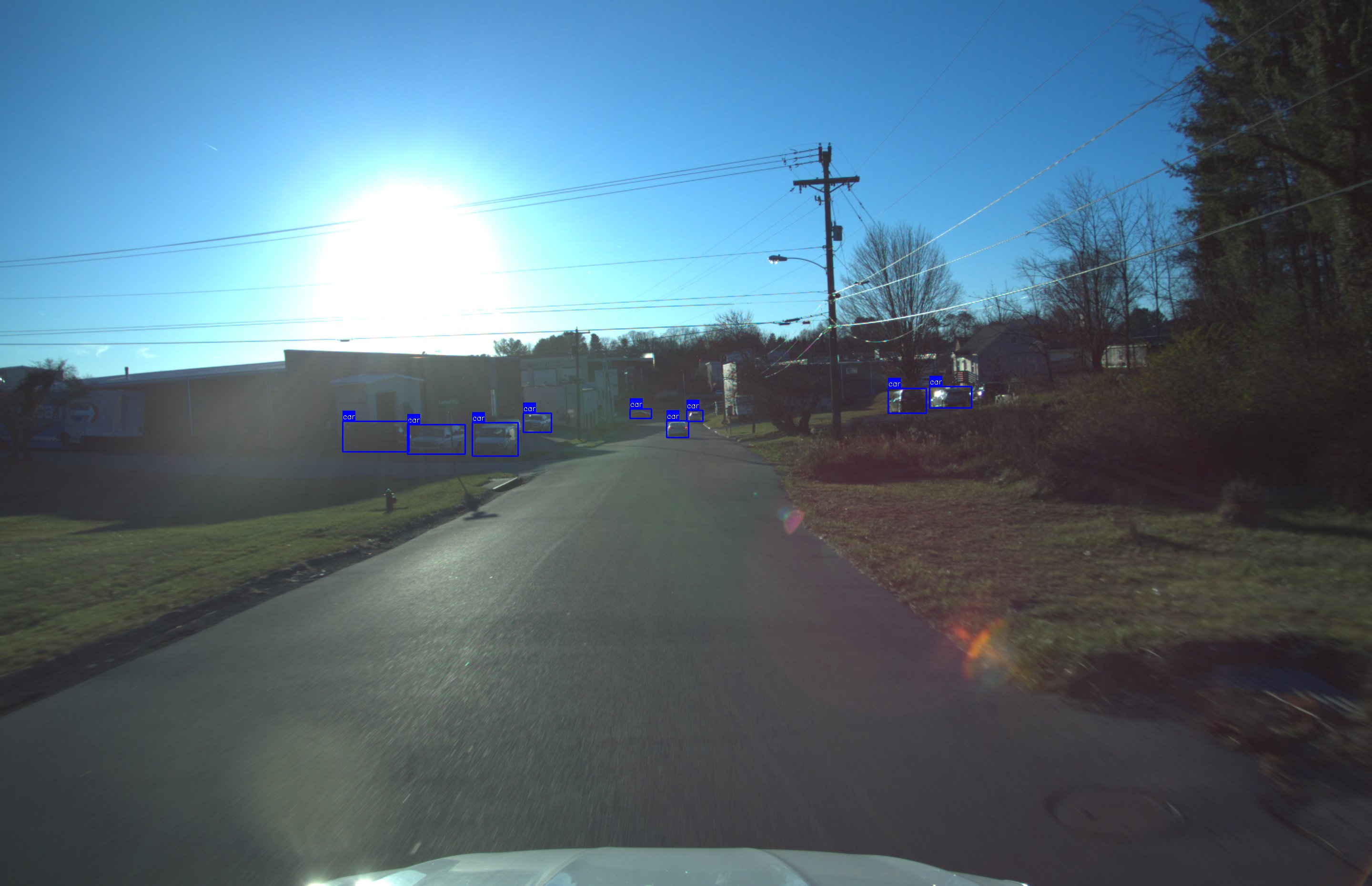} &
      \includegraphics[width=0.195\textwidth]{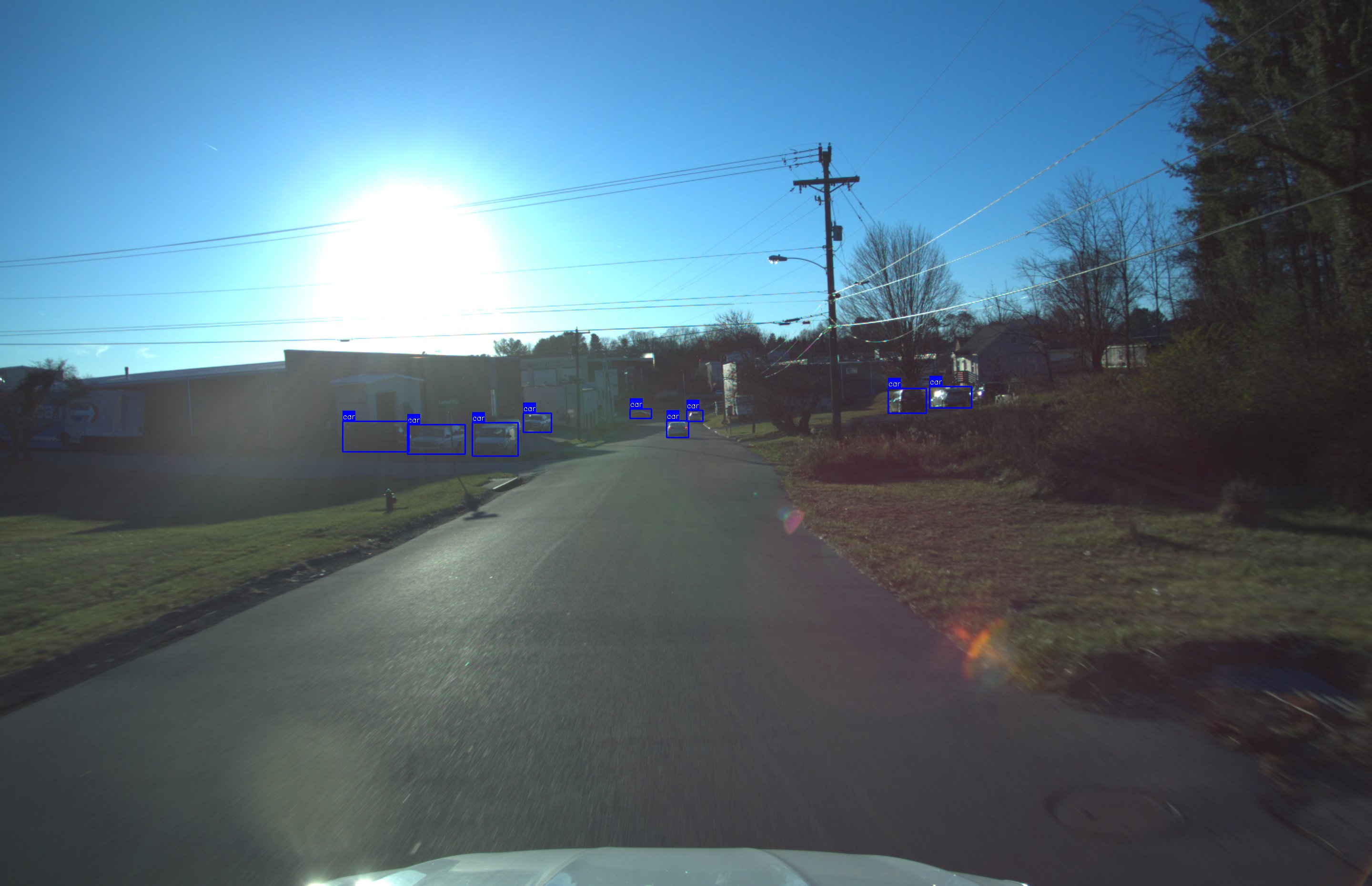} \\[0.2pt]
    
      % Row 2: Highway No Light
      \rotatebox[origin=l]{90}{\textbf{\footnotesize High. No Light}} &
      \includegraphics[width=0.195\textwidth]{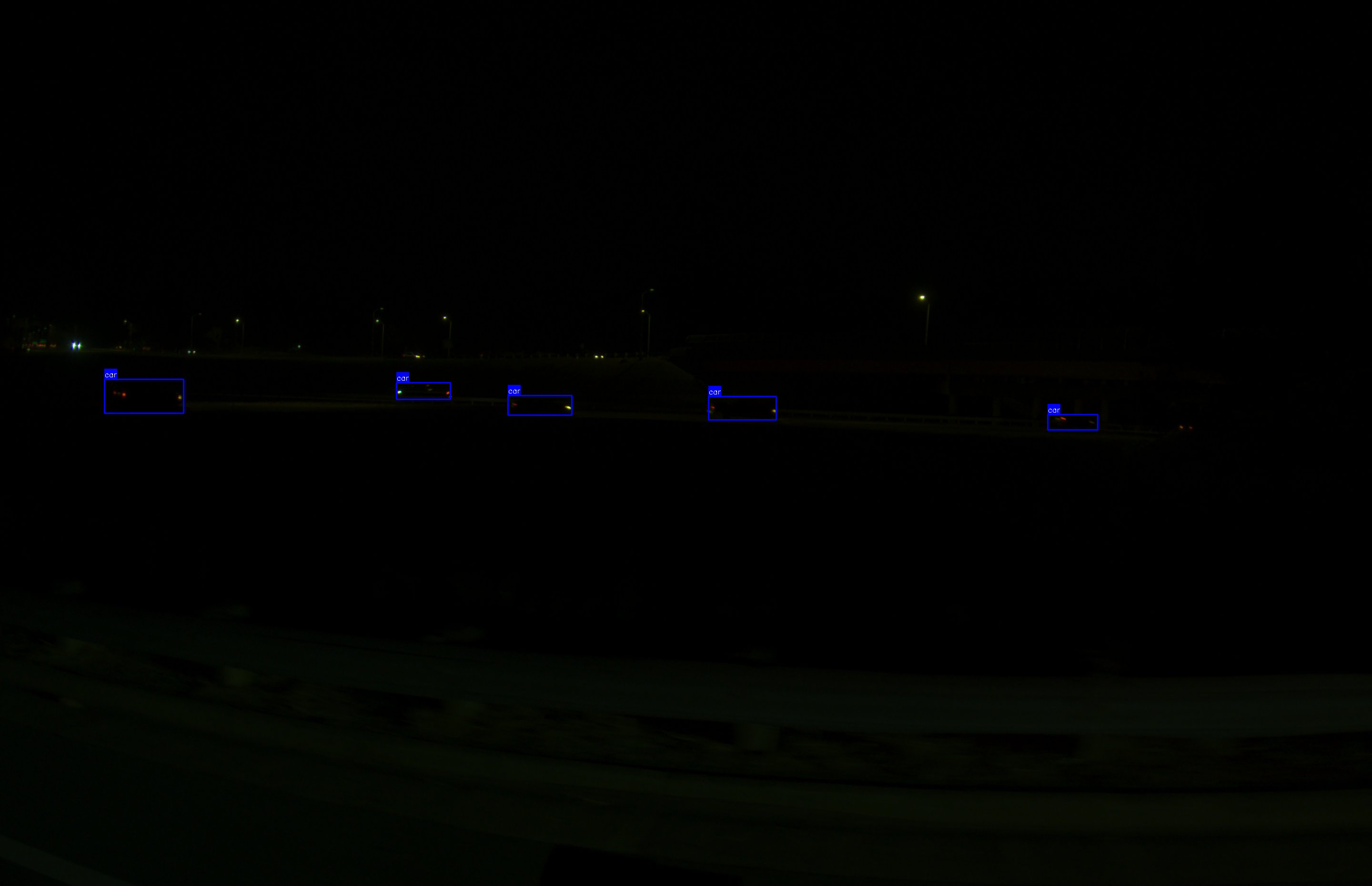} &
      \includegraphics[width=0.195\textwidth]{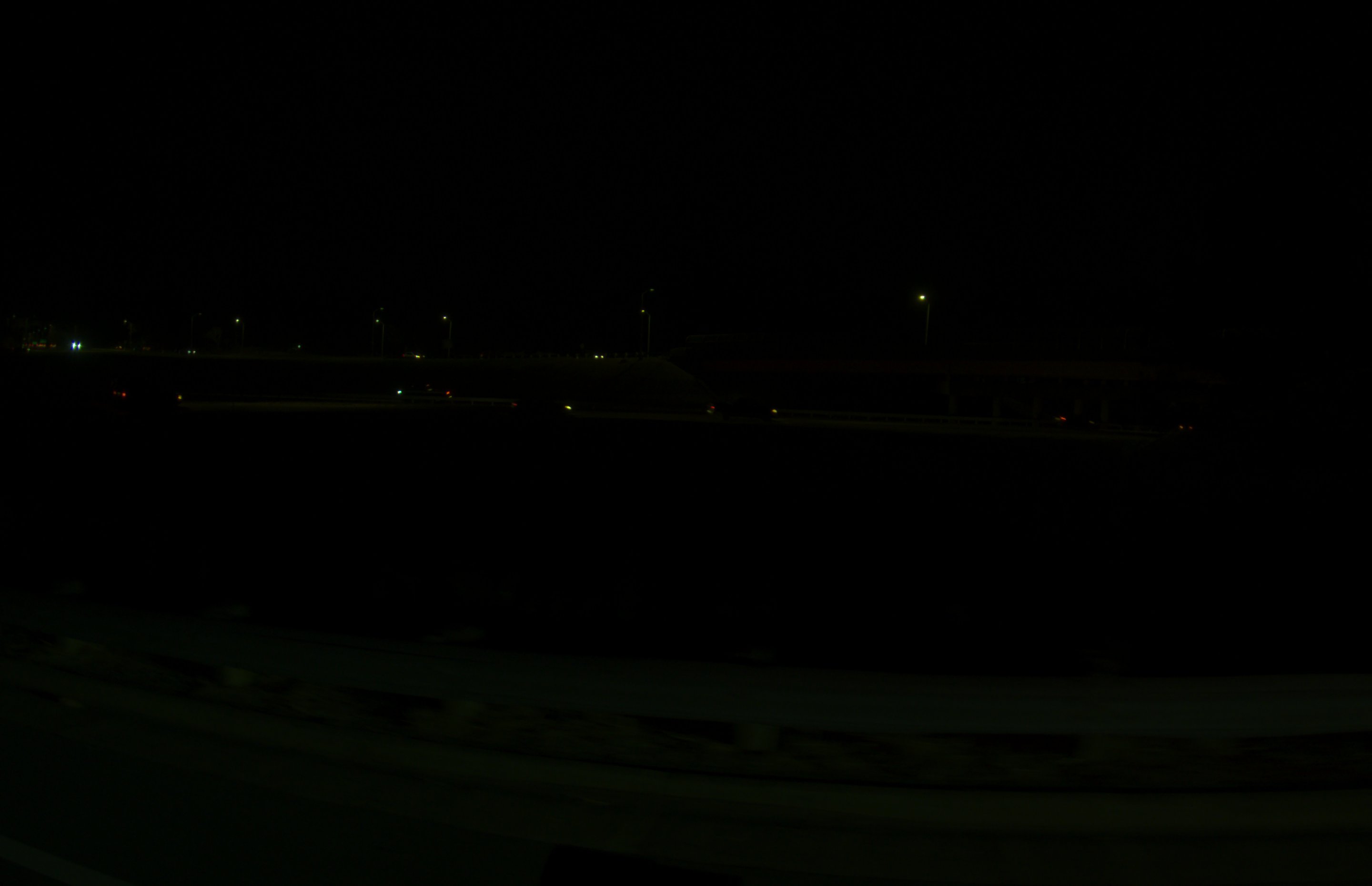} &
      \includegraphics[width=0.195\textwidth]{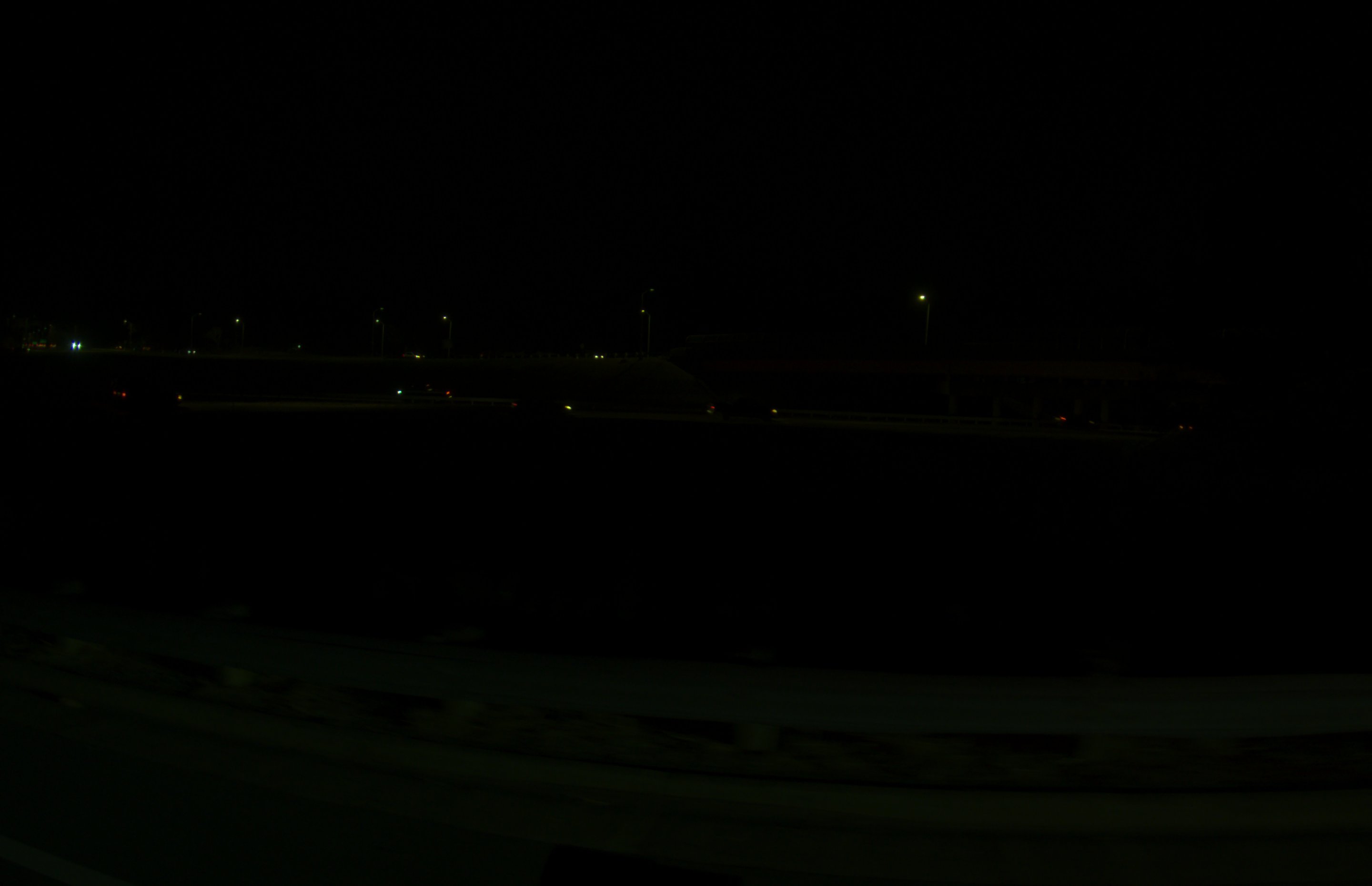} &
      \includegraphics[width=0.195\textwidth]{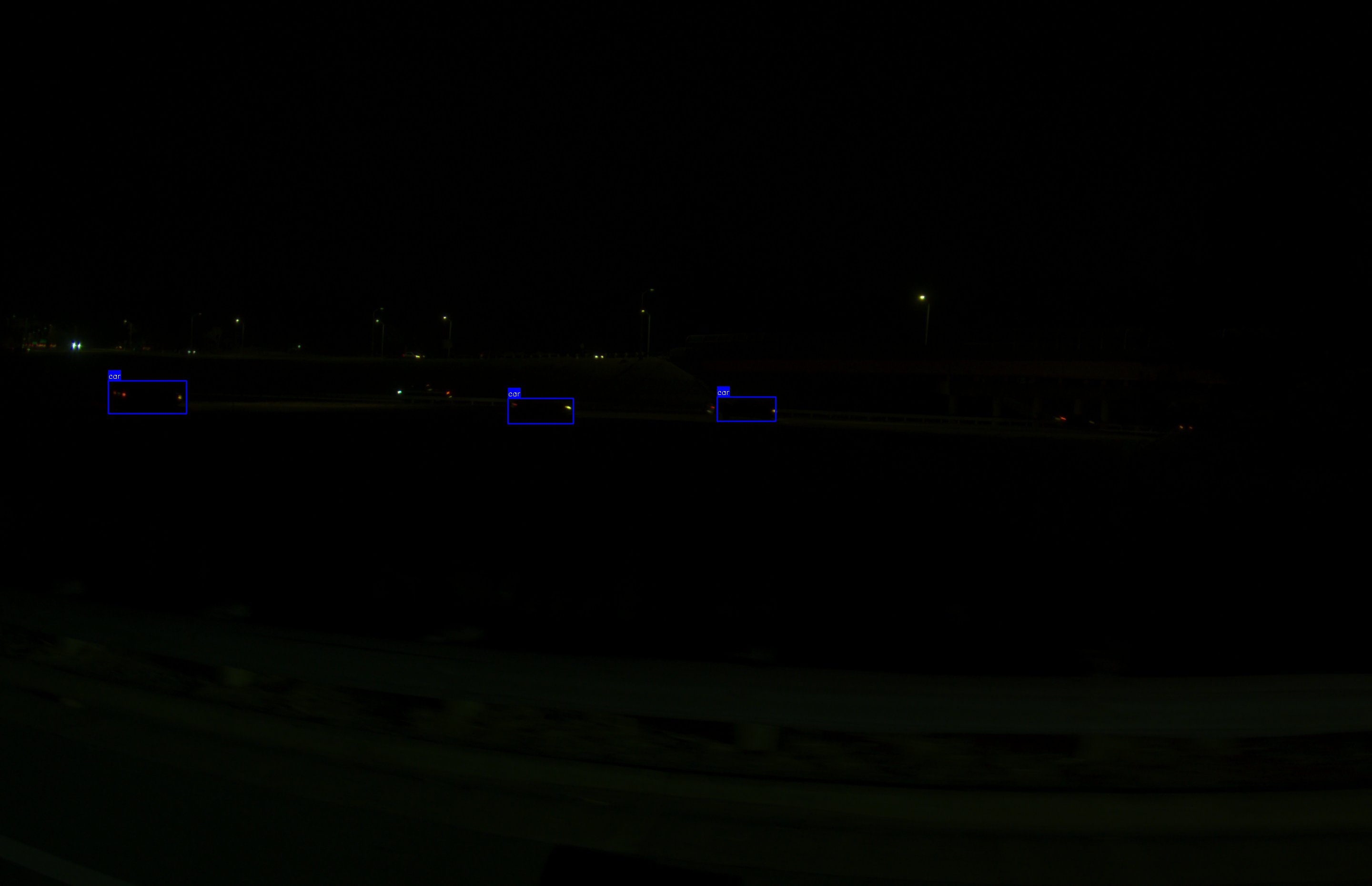} &
      \includegraphics[width=0.195\textwidth]{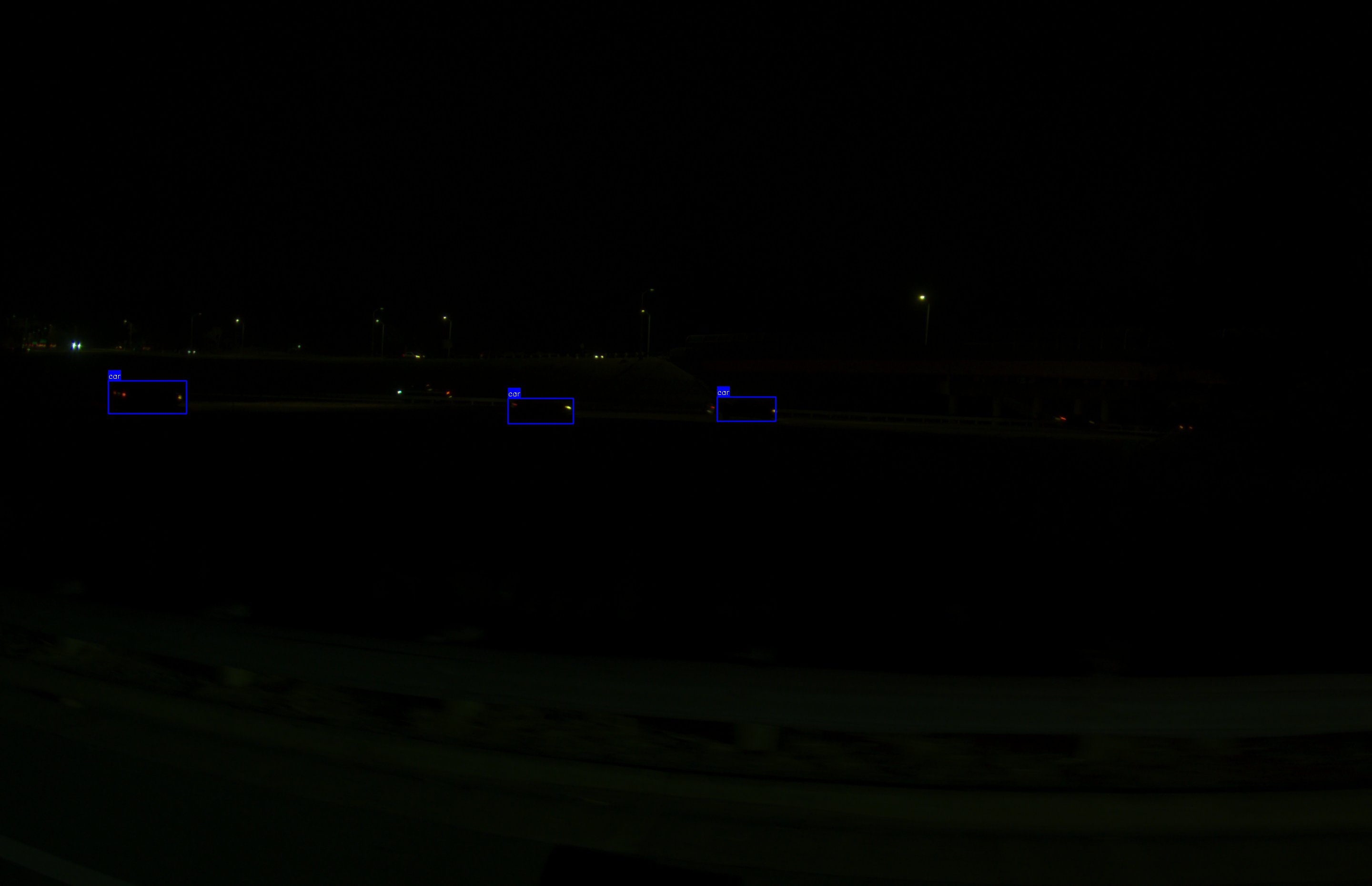} \\[0.2pt]
    
      % Row 3: Campus Fog
      \rotatebox[origin=l]{90}{\textbf{\footnotesize Campus Fog}} &
      \includegraphics[width=0.195\textwidth]{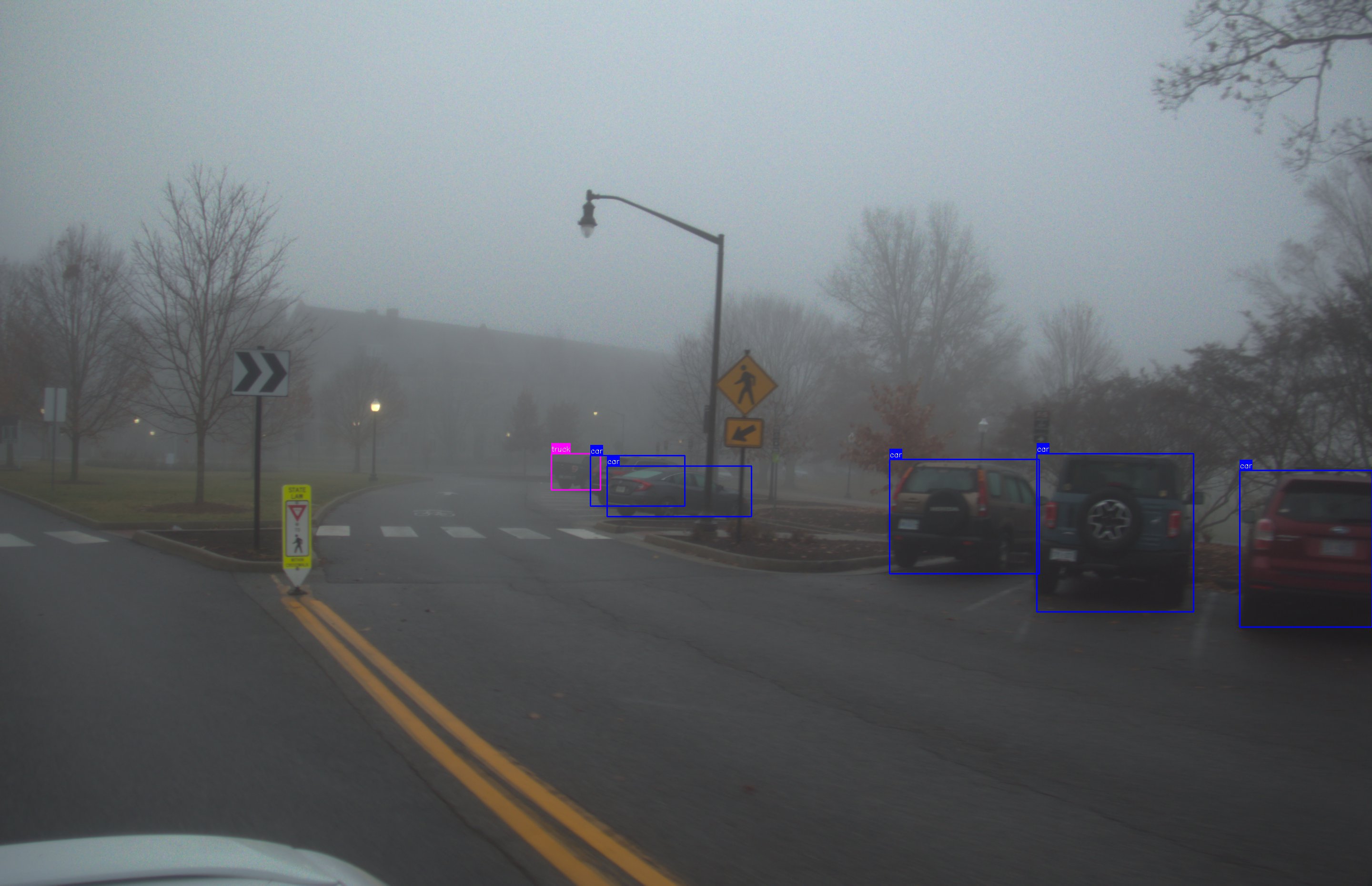} &
      \includegraphics[width=0.195\textwidth]{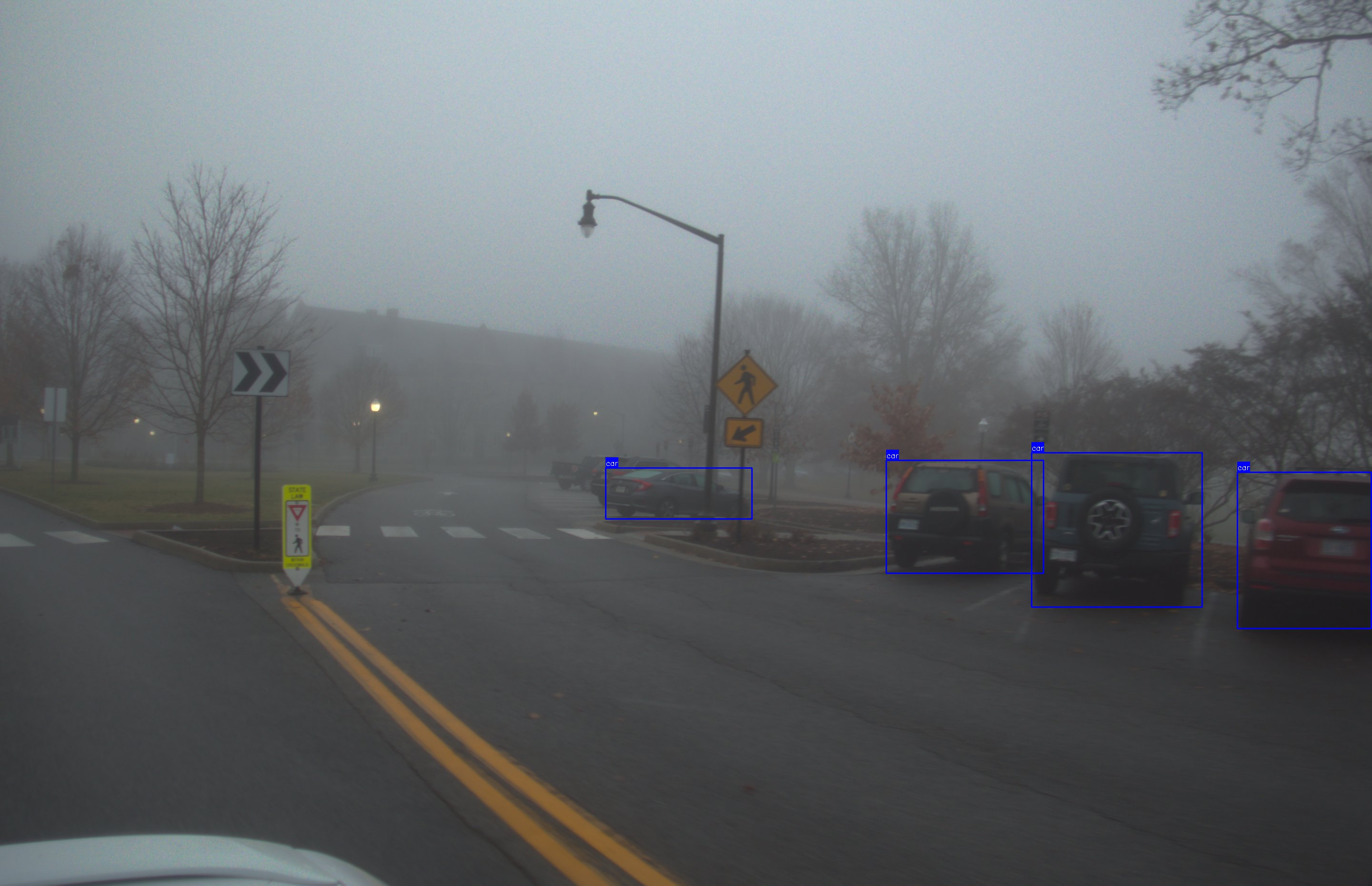} &
      \includegraphics[width=0.195\textwidth]{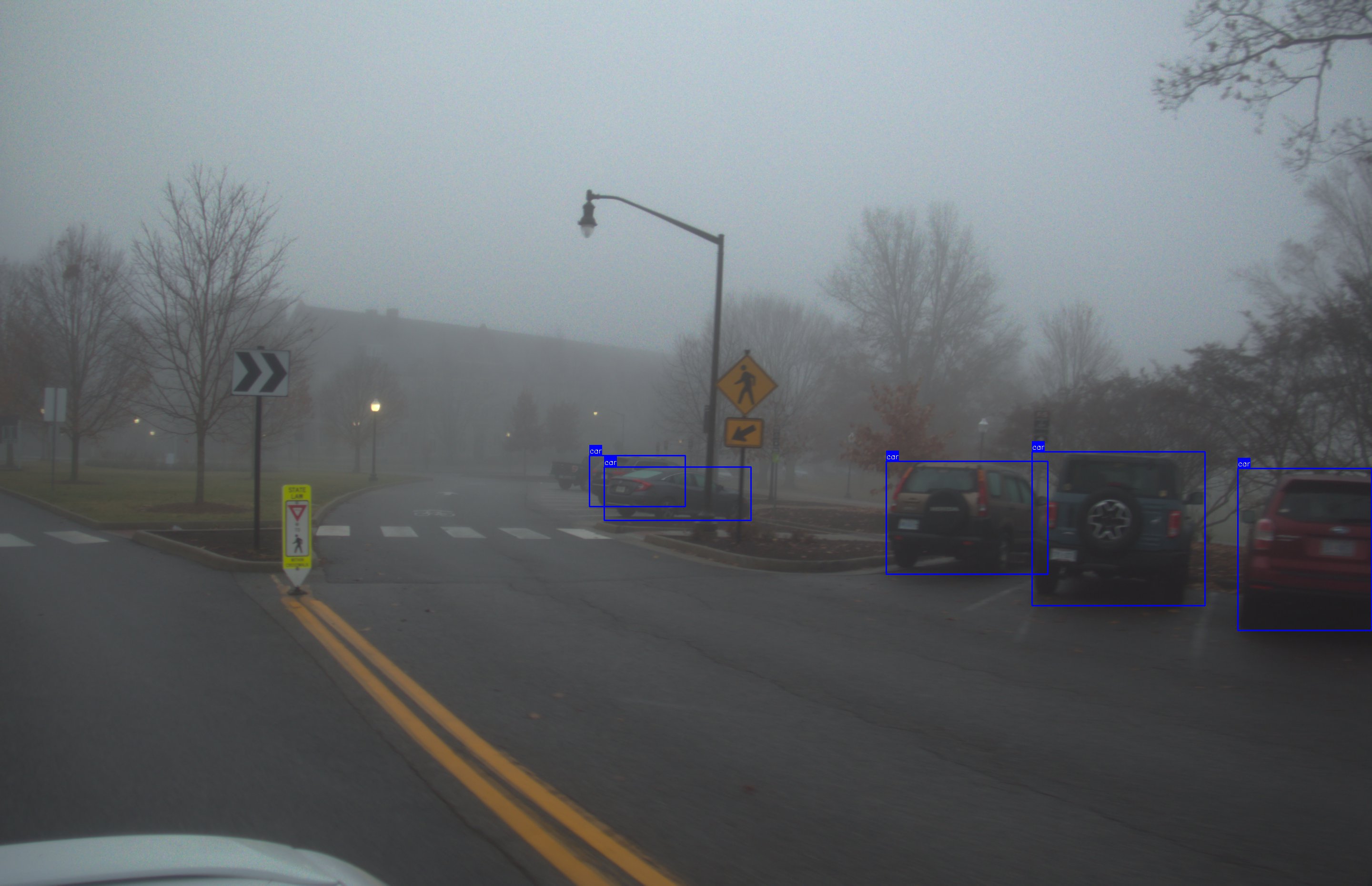} &
      \includegraphics[width=0.195\textwidth]{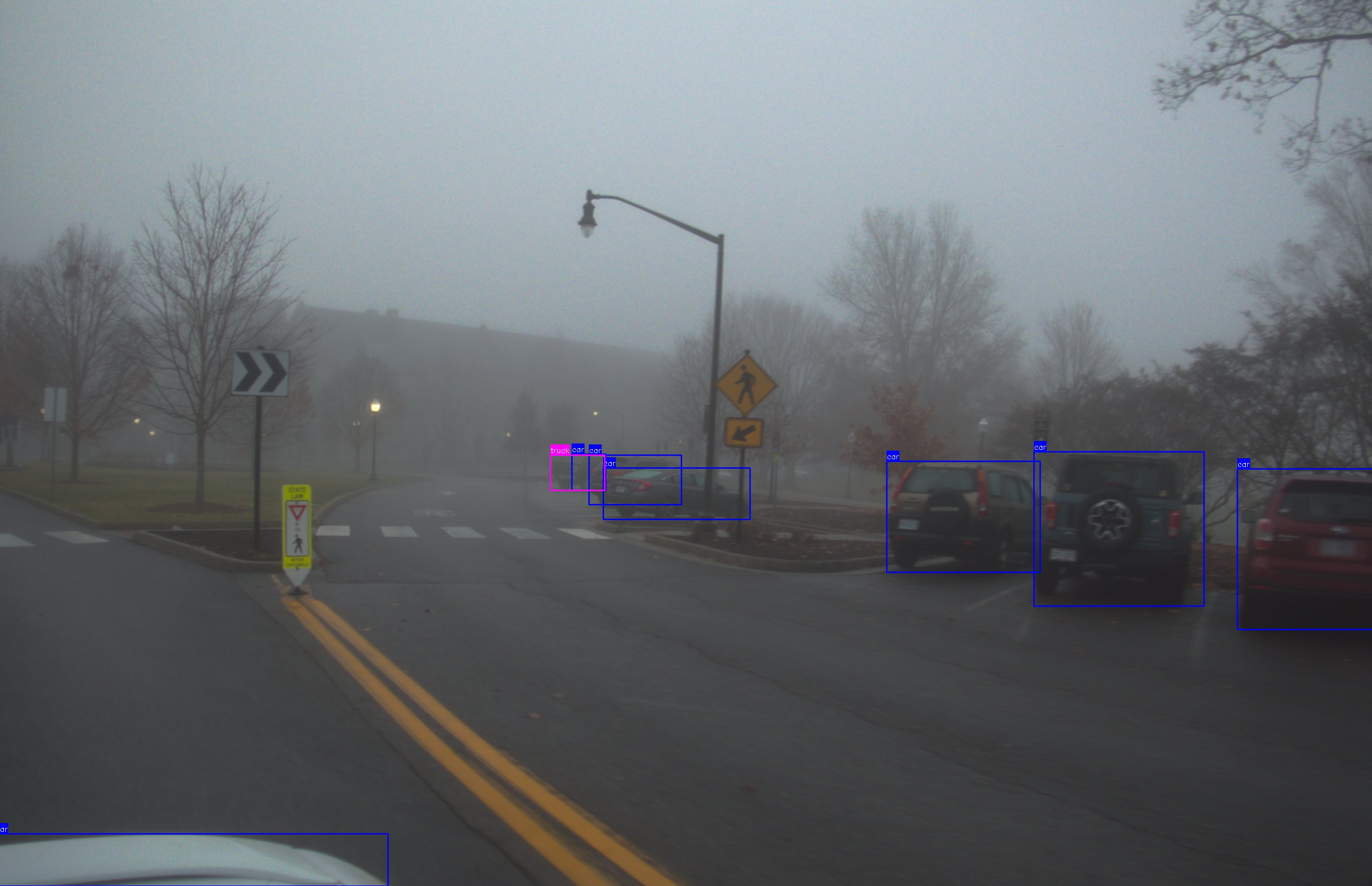} &
      \includegraphics[width=0.195\textwidth]{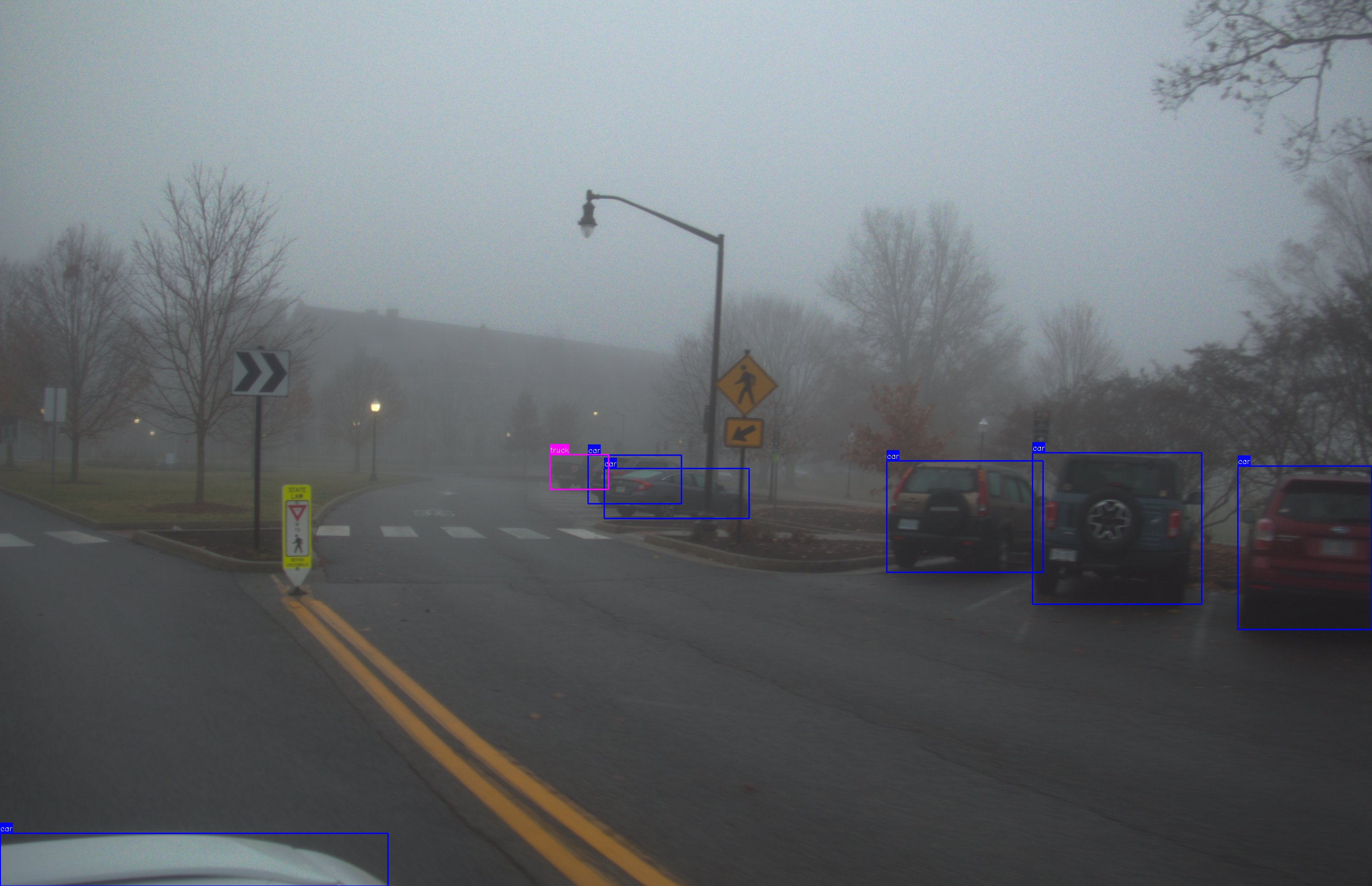} \\[0.2pt]
    
      % Row 4: Rural Rain
      \rotatebox[origin=l]{90}{\textbf{\footnotesize Rural Rain}} &
      \includegraphics[width=0.195\textwidth]{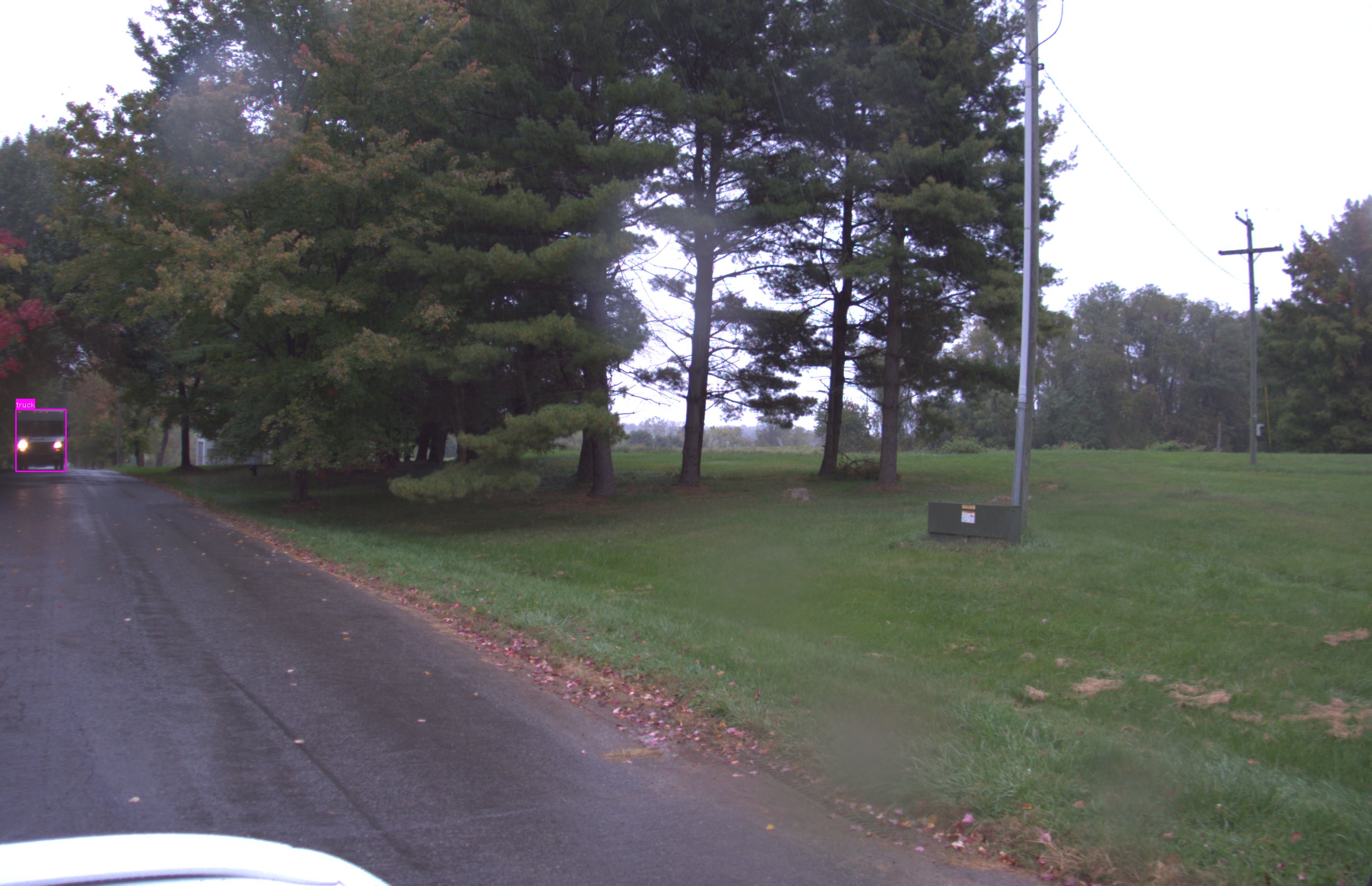} &
      \includegraphics[width=0.195\textwidth]{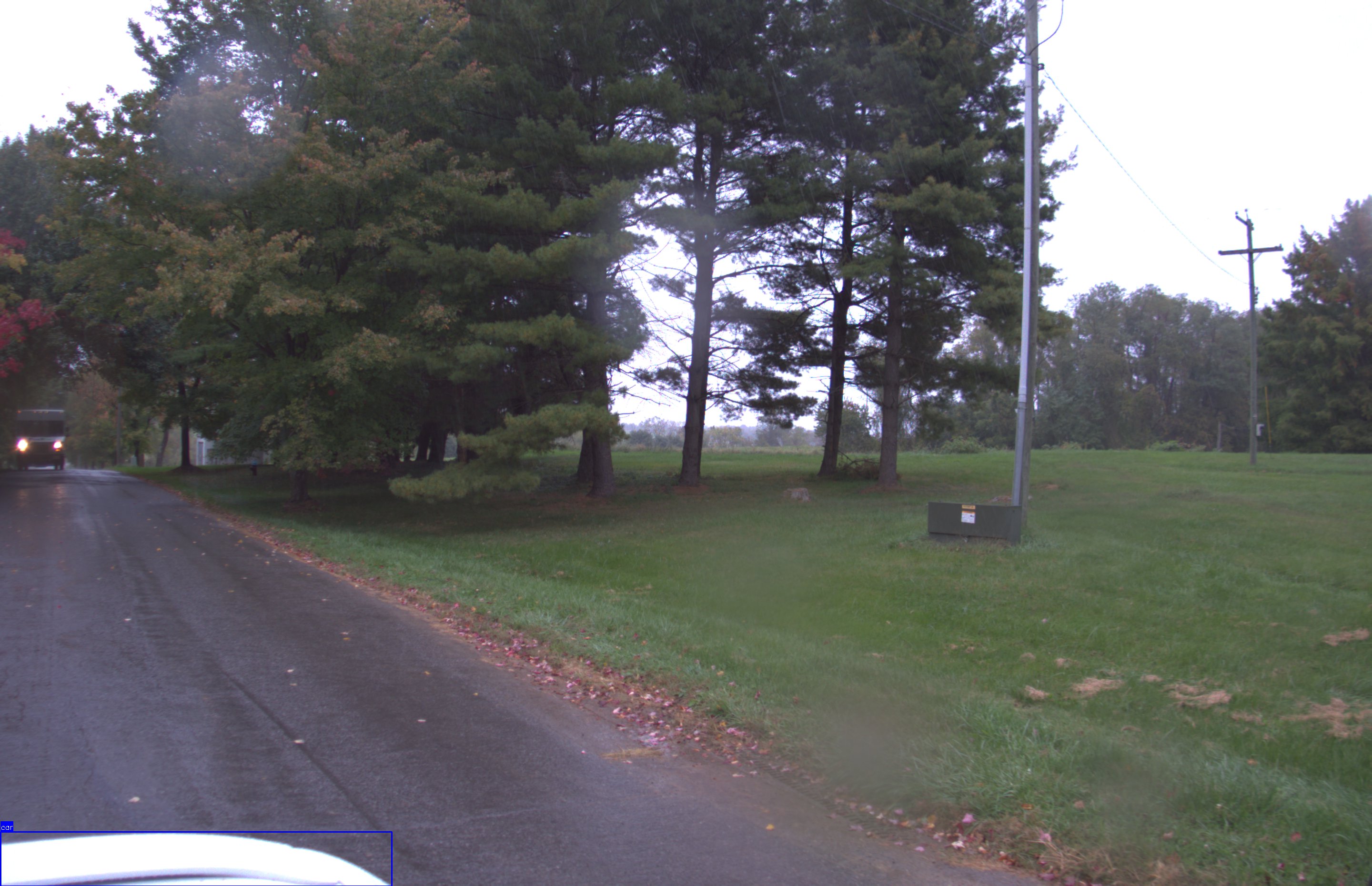} &
      \includegraphics[width=0.195\textwidth]{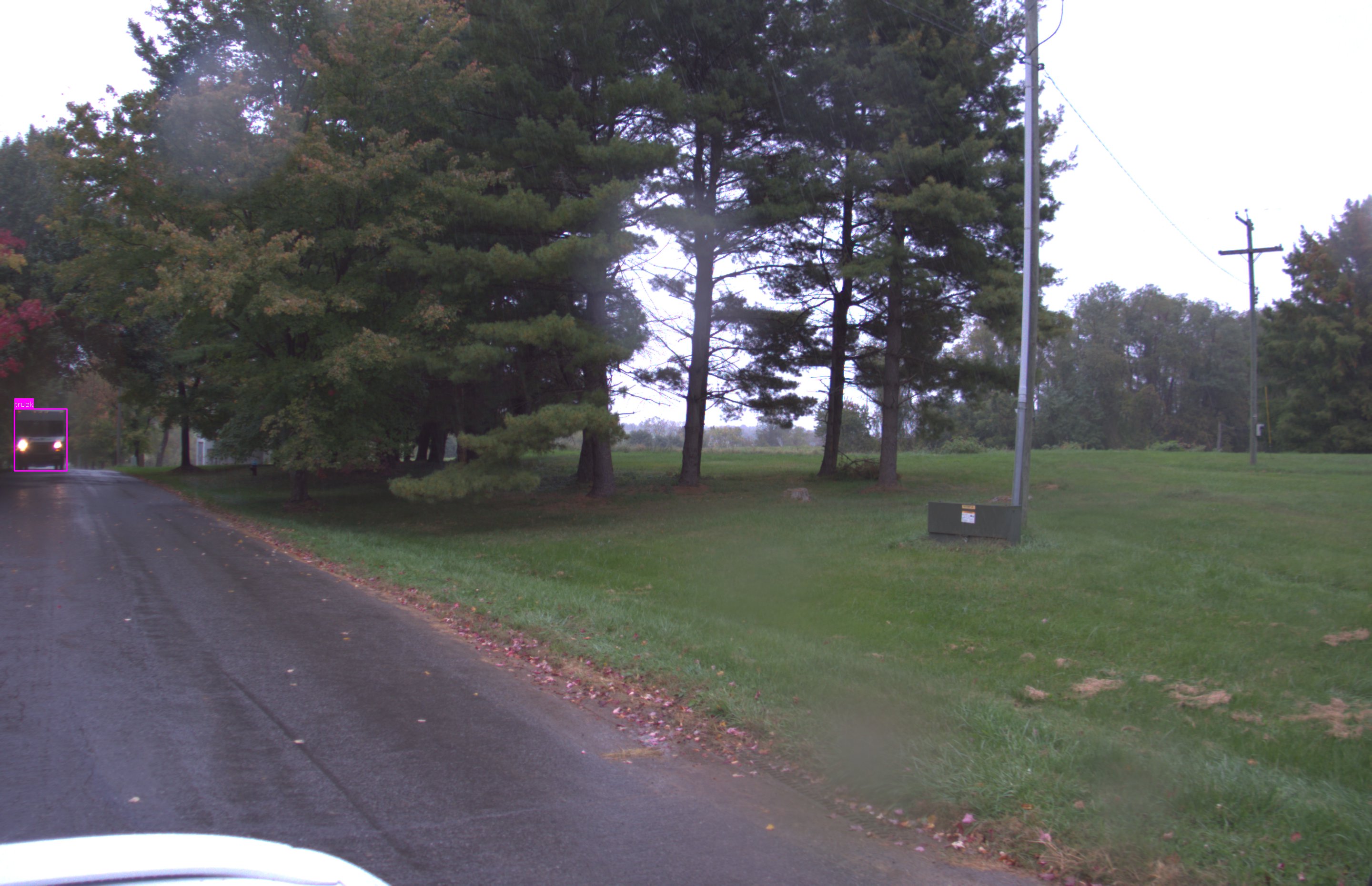} &
      \includegraphics[width=0.195\textwidth]{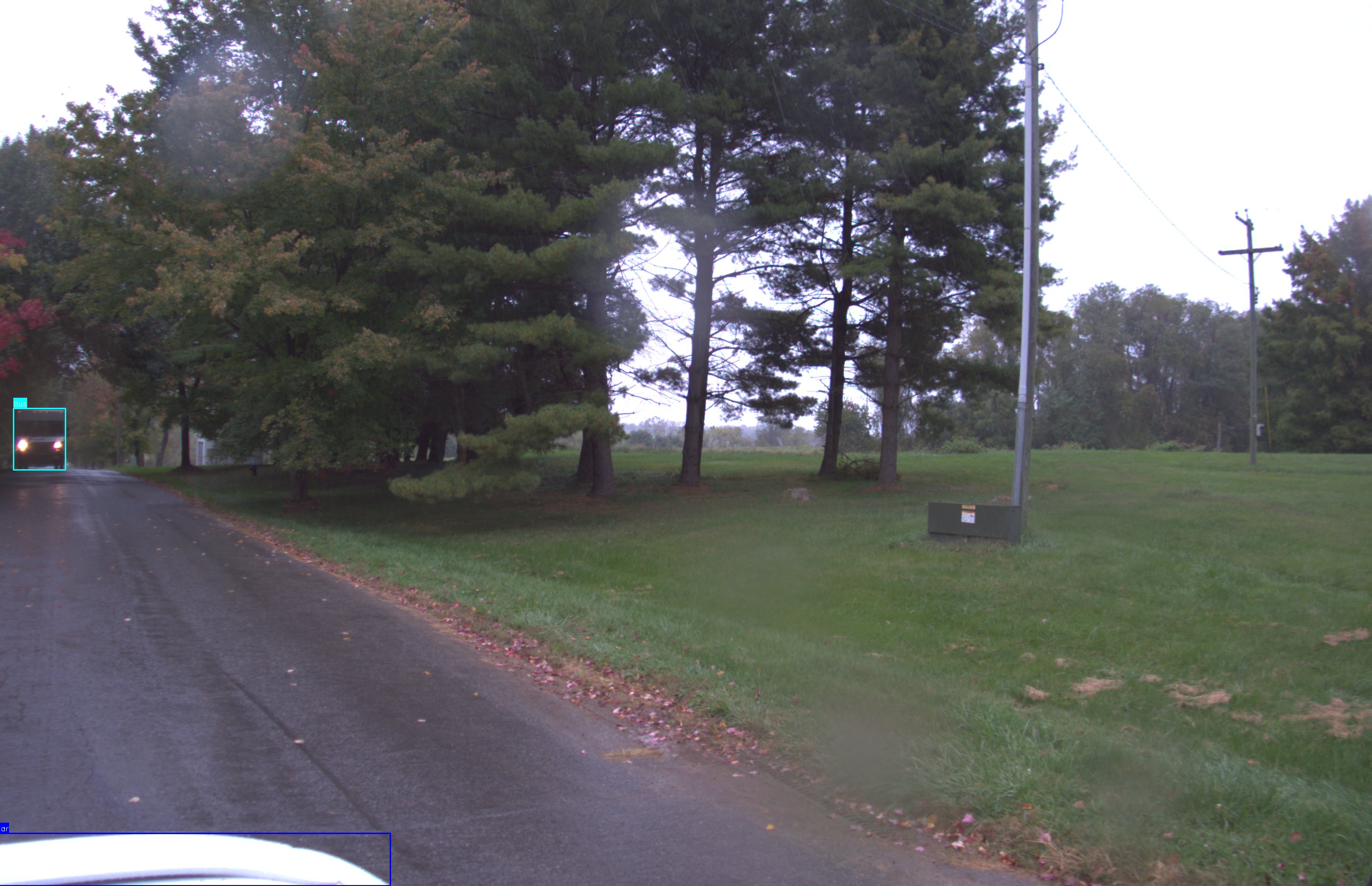} &
      \includegraphics[width=0.195\textwidth]{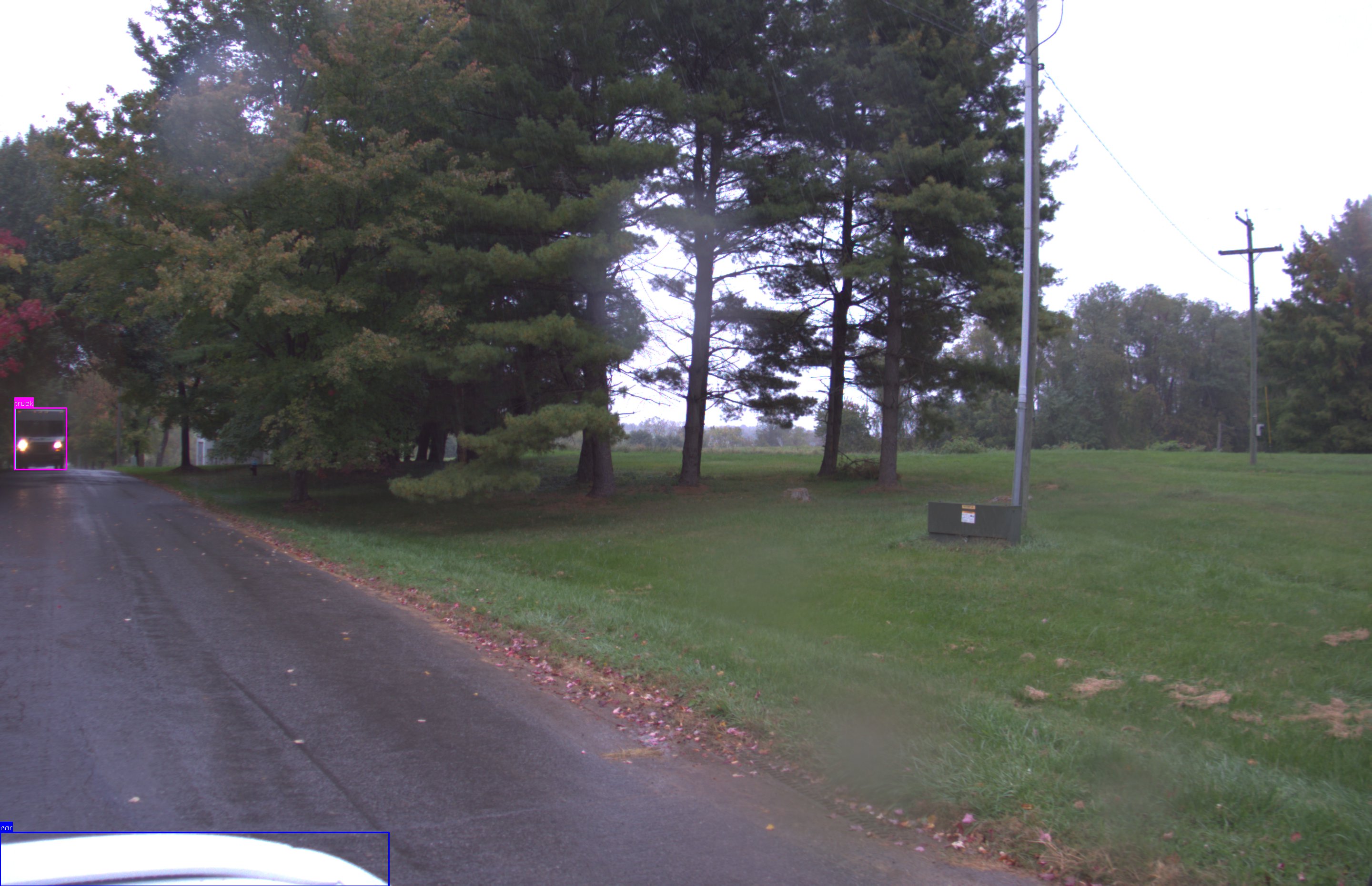} \\
    \end{tabular}
    
    \caption{Qualitative detection comparison across four operational scenarios: Residential Direct Sunlight, Highway No Light, Campus Fog, and Rural Rain. Columns display (a) Ground Truth manual annotations alongside outputs from (b) Baseline YOLOv8, (c) Co-DETR, (d) SAM3, and (e) Fine-tuned YOLOv8. Note that while baseline YOLOv8 misses heavily obscured targets, the foundation-guided fine-tuned model successfully recovers those bounding boxes (Best viewed in color and zoomed in).}
    \label{fig:qualitative_results}
    \vspace{-10pt}
\end{figure*}

To visually evaluate our fine-tuning approach, Fig.~\ref{fig:qualitative_results} compares predictions from the three candidate models and the fine-tuned detector against the ground-truth across four representative scenarios: Residential Direct Sunlight, Highway No Light, Campus Fog, and Rural Rain. We selected these specific scenes to capture pronounced instances of each condition across all four route types. Residential Direct Sunlight demonstrates our largest improvement after fine-tuning (+32.73\% mAP), whereas Highway No Light represents the lowest performance across all models. Finally, Campus Fog and Rural Rain add environmental diversity and illustrate detector behavior under distinct forms of visual degradation.

Across all four scenarios, baseline YOLOv8 performs worst among the evaluated models due to its sensitivity to environmental noise. Its detections remain restricted to well-lit, close-range targets, failing to detect occluded, distant, or poorly illuminated objects in adverse conditions like Highway No Light and Rural Rain. In contrast, the fine-tuned YOLOv8 model detects more objects across every scenario than its baseline counterpart, demonstrating a clear accuracy improvement.

Comparing predictions between Co-DETR and SAM3 confirms SAM3's greater overall robustness in adverse conditions, supporting its role as our offline auto-annotator. While Co-DETR occasionally identifies targets SAM3 misses or misclassifies, such as the truck in Rural Rain, its performance degrades in severe conditions, where it fails to detect any vehicles in Highway No Light.

Comparing fine-tuned YOLOv8 against SAM3 shows that predictions made by the fine-tuned detector closely match those of SAM3, rather than those of the baseline YOLOv8. However, the fine-tuned model does not blindly replicate SAM3's pseudo-labels. In Campus Fog, it eliminates a false-positive vehicle predicted by SAM3, and in Rural Rain, it corrects a mislabeled vehicle class. This demonstrates that fine-tuning allows YOLOv8 to absorb SAM3's detection strengths without compromising its own classification reliability.

\section{Conclusion}
\label{sec:conclusion}
In this paper, we demonstrated that standard object detectors can adapt to adverse real-world conditions without modifying their architecture or extensive training with manually annotated data. We first compared the performance of three distinct candidate models on our custom dataset covering diverse routes, weather, and lighting conditions. Due to SAM3's optimal balance of accuracy and stability across these 25 real-world driving scenarios, we selected it as our offline auto-annotator. Fine-tuning a baseline YOLOv8 model on these pseudo-labels generated by SAM3 yielded performance gains, particularly in adverse environments like Highway Fog and Residential Direct Sunlight, resulting in mAP gains of 28.65\% and 32.73\%, respectively. Overall, the fine-tuned YOLOv8 achieved a 16.04\% mAP improvement compared to the baseline model, reaching an average of 50.23\% mAP across all scenarios. However, this overall performance indicates that passive camera systems still face physical limits in the presence of real-world physical noise. Achieving true operational safety ultimately requires integrating these optimized pipelines with active multi-sensor fusion algorithms to ensure reliable perception.
\IEEEpubidadjcol

\section*{Acknowledgments}
This work builds upon the master's thesis of Xuelai Du \cite{du2024development}. The authors greatly appreciate the extensive data collection and initial experimental setup that enabled this research.
% The authors thank Xuelai Du for the extensive data collection and initial experimental setup established in their master's thesis \cite{du2024development}.

\bibliographystyle{IEEEtran}
\bibliography{references}

% \newpage
\begin{IEEEbiographynophoto}{Sepideh Gohari}
received the B.S. degree in electrical engineering and the M.S. degree in artificial intelligence and robotics from Ferdowsi University of Mashhad, Mashhad, Iran. She is currently pursuing the Ph.D. degree in electrical and computer engineering at the Autonomous Robots and Vehicles Lab (ARVL), at Virginia Commonwealth University (VCU), Richmond, VA, USA. Her research interests include computer vision, autonomous driving, robotics, and artificial intelligence.
\end{IEEEbiographynophoto}

\begin{IEEEbiographynophoto}{Goodarz Mehr}
received the B.Sc. degree in mechanical engineering from Sharif University of Technology, Tehran, Iran, in 2016 and the M.Sc. and Ph.D. degrees in mechanical engineering from Virginia Tech, Blacksburg, VA, USA, in 2023 and 2024, respectively. He is currently a postdoctoral research associate at Autonomous Robots and Vehicles Lab (ARVL) at Virginia Commonwealth University (VCU), Richmond, VA, USA. His research interests include multi-agent robotics, stochastic planning models, and cooperative perception.
\end{IEEEbiographynophoto}

\begin{IEEEbiographynophoto}{Azim Eskandarian}
(IEEE Fellow) received the B.S. and D.Sc. degrees in mechanical engineering from George Washington University (GWU), Washington, D.C., USA, and the M.S. degree in mechanical engineering from Virginia Tech, Blacksburg, VA, USA.

He has been the Alice T. and William H. Goodwin Jr. Dean of the College of Engineering at Virginia Commonwealth University (VCU), Richmond, USA, since 2023, where he also established the Autonomous Robots and Vehicle Laboratory (ARVL). Before that, he was a Professor and the Head of the Department of Mechanical Engineering at Virginia Tech since 2015, where he became the Nicholas and Rebecca Des Champs Chair Professor in 2018, and where he established the Autonomous Systems and the Intelligent Machines Laboratory, to conduct research in intelligent and autonomous vehicles and mobile robotics. As a Professor at George Washington University, he was the Co-Founder of the National Crash Analysis Center in 1992, the founding director of the Center for Intelligent Systems Research from 1995 to 2015, and the director of the University's area of excellence in Transportation Safety and Security from 2003 to 2015. He was an Assistant Professor at The Pennsylvania State University in York, PA, USA, from 1989 to 1992 and an Engineer/Project Manager in the industry from 1983 to 1989.  

Dr. Eskandarian is a Fellow of IEEE, ASME, and SAE. He was elected to the Virginia Academy of Science, Engineering, and Medicine (VASEM) in 2026. He received the IEEE Intelligent Transportation Society Outstanding Researcher Award in 2017 and the GWU School of Engineering Outstanding Researcher Award in 2013. He served as Editor-in-Chief of the IEEE Transactions on Intelligent Transportation Systems from 2019 to 2023.
\end{IEEEbiographynophoto}

\end{document}